\documentclass[sigconf,nonacm]{acmart}

\AtBeginDocument{%
  \providecommand\BibTeX{{%
    \normalfont B\kern-0.5em{\scshape i\kern-0.25em b}\kern-0.8em\TeX}}}

\usepackage{mathrsfs}
\usepackage{CJKutf8}
\usepackage[outdir=./]{epstopdf}
\usepackage{url}
\usepackage{makecell}
\usepackage{xcolor}
\usepackage{multirow}
\usepackage{graphicx}
\usepackage[normalem]{ulem}
\usepackage{booktabs}
\useunder{\uline}{\ul}{}

\usepackage{bbm}
\usepackage{amsmath,amsfonts}

\usepackage{amssymb}
\usepackage{pifont}% http://ctan.org/pkg/pifont
\usepackage{caption}
\usepackage{enumitem}
\newcommand{\note}[1]{{\color{blue}{[\textbf{attention}: #1]}}}

\begin{document}
\begin{CJK}{UTF8}{gbsn}

\title[]{Who Should Review Your Proposal? Interdisciplinary Topic Path Detection for Research Proposals}

%\title{How to Assign Your Proposal for Review? Interdisciplinary Topic Path Detection for Research Proposals}

% \title{Who Should Review your Proposal?
% An Interdisciplinary Research Proposal Classification Method with the Right Granularity}

\author{Meng Xiao$^{1,2}$, Ziyue Qiao$^{1,2}$, Yanjie Fu$^{3}$,Hao Dong$^{1,2}$, Yi Du$^{1,*}$,Pengyang Wang$^4$ \\Dong Li$^5$, Yuanchun Zhou$^{1}$}
\renewcommand{\shortauthors}{ }
\thanks{$^*$Corresponding author.}
\affiliation{%
\institution{\text{$^1$Computer Network Information Center, Chinese Academy of Sciences, Beijing},
\text{$^2$University of Chinese Academy of Sciences, Beijing},
\text{$^3$Department of Computer Science, University of Central Florida, Orlando},
\text{$^4$University of Macau, Macau}
\text{$^5$National Natural Science Foundation of China, Information Center, Beijing}
}
\country{$^{1,2}$\{shaow, qiaoziyue,donghao\}@cnic.cn, $^1$\{duyi, zyc\}@cnic.cn, $^3$yanjie.fu@ucf.edu,$^4$pywang@um.edu.mo,\\$^5$lidong@nsfc.gov.cn}
}
\begin{abstract}
%Yanjie
The peer merit review of research proposals has been the major mechanism to decide grant awards. 
Nowadays, research proposals have become increasingly interdisciplinary. It has been a longstanding challenge to assign proposals to appropriate reviewers. One of the critical steps in reviewer assignment is to generate accurate interdisciplinary topic labels for proposals. Existing systems mainly collect topic labels manually reported by discipline investigators. However, such human-reported labels can be non-accurate and incomplete. What role can AI play in developing a fair and precise proposal review system? In this evidential study, we 
collaborate with the National Science Foundation of China to address the task of automated interdisciplinary topic path detection. For this purpose, we develop a deep Hierarchical Interdisciplinary Research Proposal Classification Network (HIRPCN). We first propose a hierarchical transformer to extract the textual semantic information of proposals. 
We then design an interdisciplinary graph and leverage GNNs to learn representations of each discipline in order to extract interdisciplinary knowledge. 
After extracting the semantic and interdisciplinary knowledge, we design a level-wise prediction component to fuse the two types of knowledge representations and detect interdisciplinary topic paths for each proposal. 
We conduct extensive experiments and expert evaluations on three real-world datasets to demonstrate the effectiveness of our proposed model.

\end{abstract}


\maketitle

\section{Introduction}\label{sec:introduction}
% \subsection{Background}
% introduction
% 问题描述 
% 几个关键点让这个问题必须要做
% - 1 这样一个系统能够advance effectiveness and fairness
% - 2 数据量增加导致不得不引入ai辅助
% 观察数据后数据的特点以及三个点
%  - 学科体系的结构
%  - 交叉学科的涌现
%  - 申请书具有多文本特性，在现实的评审过程中，这些特性有着重要的意义
% 模型部分
% 简要介绍
% 贡献
% 接下来部分的介绍
% As scientific knowledge in a wide range of disciplines has advanced, scholarshave become increasingly aware of the need to link disciplinary fields to morefully answer critical questions, or to facilitate application of knowledge in aspecific area. For example, the discovery that tobacco use was associated withhigh rates of lung disease was not sufficient to lead to smoking cessation; theaddition of research on risk assessment, motivation, and reasoned action wereall important in designing programs that have fostered the current lower ratesof tobacco use. This recognition has stimulated a steadily growing interestwithin the scientific community in developing new knowledge through re-search that combines the skills and perspectives of multiple disciplines. Thismay be in part a parallel of the wider societal interest in holistic perspectivesthat do not reduce human experience to a single dimension of descriptors, andto awareness that a number of extremely important and productive fields ofstudy are themselves interdisciplinary: biochemistry, biophysics, social psy-chology, geophysics, informatics, and others.
\begin{figure}
  \centering
  \captionsetup{justification=centering}
  \includegraphics[width=0.45\textwidth]{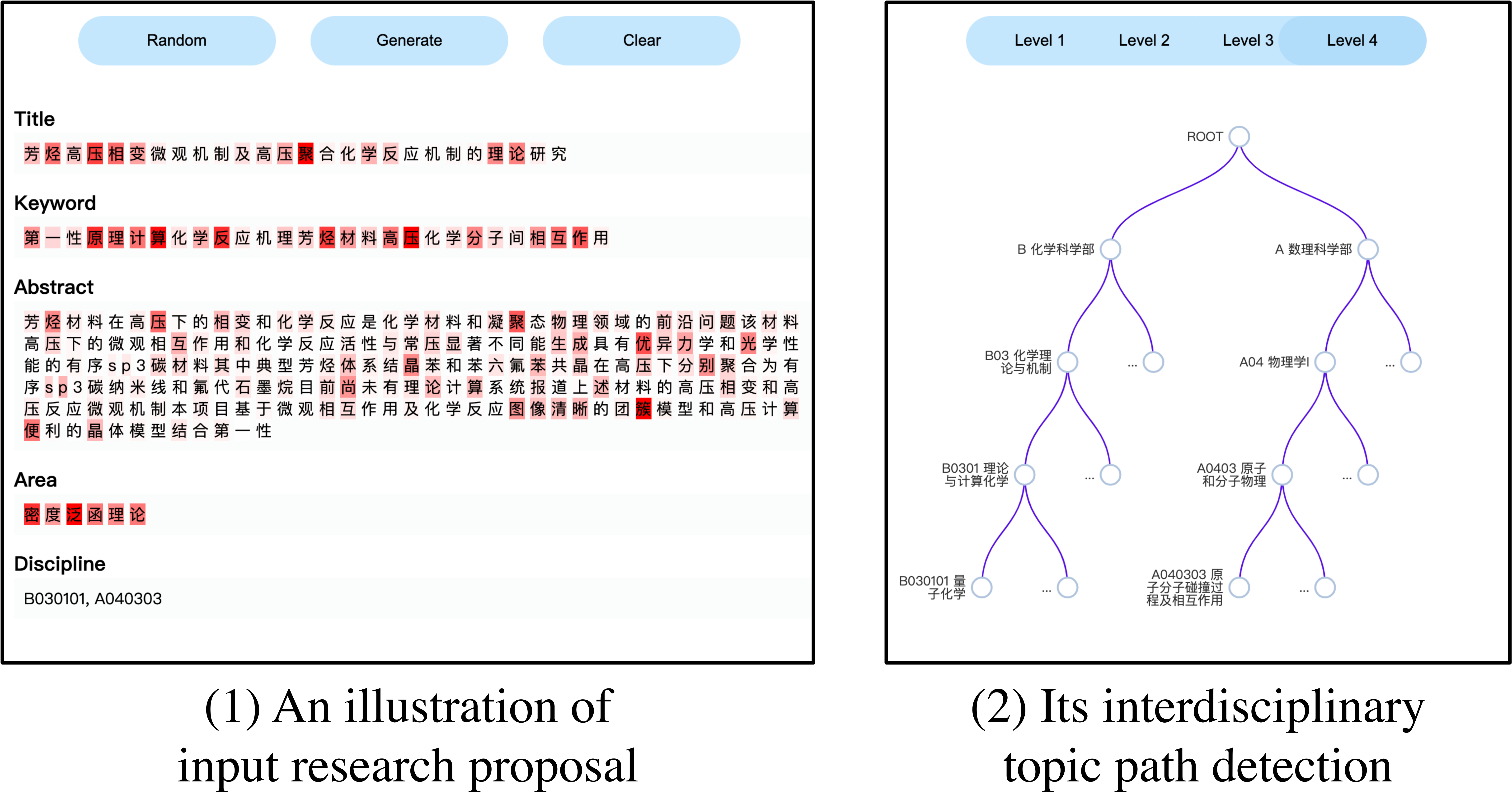}
  \vspace{-0.3cm}
  \caption{A demo of HIRPCN to aid the online interdisciplinary research proposal classification.}
  \vspace{-0.3cm}
  \label{Fig.demo}
\end{figure}
\begin{figure*}
  \centering
  \captionsetup{justification=centering}
  \includegraphics[width=0.85\textwidth]{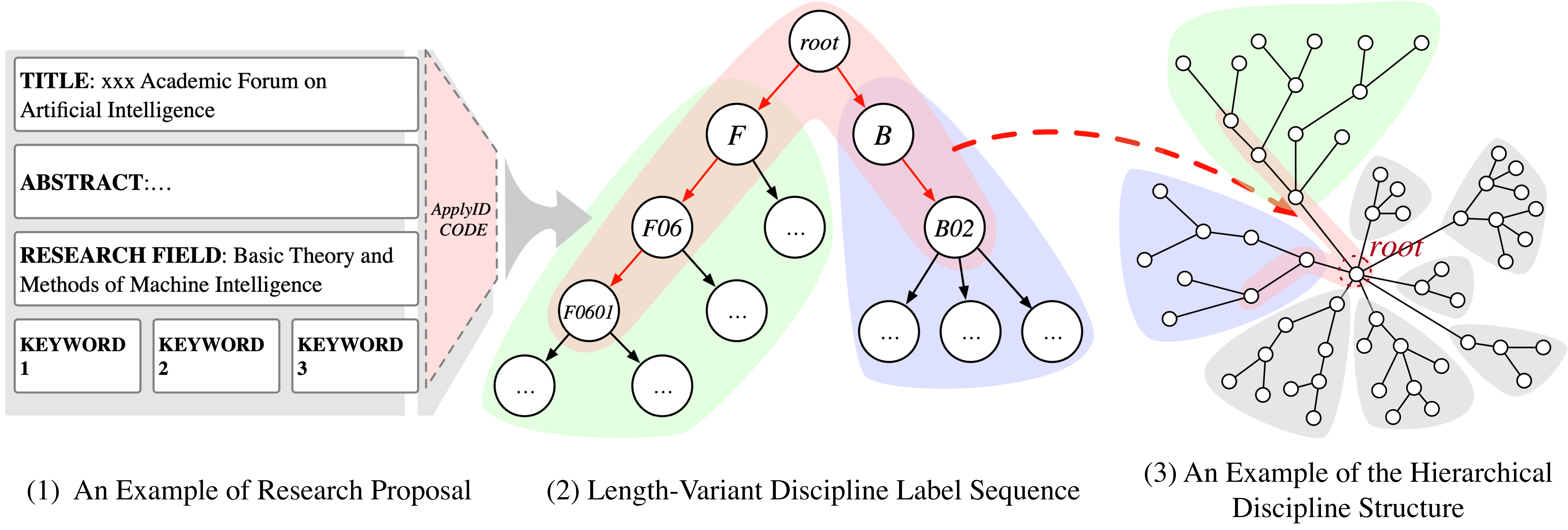}
  \vspace{-0.3cm}
  \caption{A toy model of the Interdisciplinary Research Proposal Classification.}
  \vspace{-0.3cm}
  \label{Fig.label_1}
\end{figure*}
Currently, research funding is awarded based on proposals' intellectual, educational, and socio-societal merits. Therefore, scientists submit proposals for their research to open-court competitive programs managed by government agencies (e.g., NSF). Then, proposals are assigned to appropriate reviewers to solicit review comments and ratings. 
One of the pains of running such a peer-review system is to assign the research proposal to a proper domain, thus allocating a set of domain-related reviewers to advance the review process's effectiveness and fairness. 
However, with the increasing number of universities, faculty members, and graduate student hiring, the publications, the ideas, and the submissions of research proposals have been exploding~\cite{milojevic2015, fortunato2018science}. 
For example, the Natural Science Foundation of China (NSFC) needs to process more than one million research proposals and find suitable review experts for these proposals every year. Due to this growth momentum, There is an urgent need to bring in AI to assist with proposal categorization, review assignments, panel discussions. 

Further, an increasing number of research proposals exhibit an interdisciplinary characteristic, resulting in difficulty in choosing qualified reviewers from one research field. 
For that, it is critical to classify a proposal into one or more disciplines, or we call it the Interdisciplinary Research Proposal Classification (IRPC) task. Figure \ref{Fig.label_1} shows a toy model of the IRPC task.
After analyzing large-scale proposal data from Chinese scientists' research proposals, we identify three unique data characteristics. These unique properties provide great potential to overcome the challenge in IRPC task:

(\textbf{P1}) The research proposal consists of \textbf{multi-type} of textual data, and different types of text have a distinct meaning. 
An expert will easily identify the major discipline of a proposal by reading the title, but for a fine-grainer category, one has to consider the content in each type. 
Meanwhile, since the texts in most research proposals \textbf{vary widely in length and structure}, stitching each text together results in a severe loss of information.

(\textbf{P2}) The domain knowledge, peer-reviewers, and the proposal categories are natural to organize into a discipline system. In NSFC, the disciplines category as a taxonomy system, which we call ApplyID\footnote{\url{http://www.nsfc.gov.cn/publish/portal0/tab550/}}. As shown in Figure~\ref{Fig.label_1}(3), this system contains thousands of disciplines and sub-disciplines with different levels of granularity, which exhibits a \textbf{hierarchical structure}.
Every ApplyID code prefix by a capital letter from \textit{A} to \textit{H}, representing eight major disciplines, followed by zero- to six-digit. Every two-digit number represents a sub-discipline division in a particular granularity. 
For example, in Figure~\ref{Fig.label_1}(2), the \textit{F} refers to the major discipline \textit{Information Sciences}. the \textit{F06} represents \textit{Artificial Intelligence}, a sub-discipline of \textit{Information Sciences}, and the \textit{F0601} represents \textit{Fundamentals of Artificial Intelligence}, a sub-discipline of \textit{F06}.

(\textbf{P3}) Due to the evolution of sciences, the knowledge involved in solving a particular problem has all over the range that a single domain or industry can handle~\cite{foster2015tradition}. To evaluate these proposals more fairly, find matching experts based on their related domain, and avoid the destruction of valuable pioneering and interdisciplinary research, it is essential to identify \textbf{multiple disciplines} for those proposals. 
Figure~\ref{Fig.label_1}(2) shows a interdiscipline. The example proposal is related to the research field in ApplyID \textit{F0601} and \textit{B02}. 

Based on the discussion above, we construct the discipline system and model the task as a top-down iterative process along the discipline hierarchy. For that, we propose Hierarchical Interdisciplinary Research Proposal Classification Network (HIRPCN), a deep learning model with a hierarchical multi-label classification scheme. In HIRPCN architecture, we pursue three unique strategies: First, the model utilizes the historical prediction result in every step to start prediction from any given labels. Second, we use type-token and hierarchical Transformer architecture to preserve type-specific semantic information to handle multiple textual data types. Third, we develop an Interdisciplinary Graph to represent interdisciplinarity and use its topology structure to incorporate the knowledge in each iteration step.
In conclusion, our contributions are as follows: 

% 1. 我们研究了交叉学科申请书分类中的关键问题，并
% 1. We identify the key problems of assigning proposals to related areas for review. We solve these problems via detecting topic paths on the hierarchical discipline tree by using the proposed text.

% 2. We propose HIRPCN, a hierarchical label generation framework for classifying research proposals while can generating multiple topic paths for interdisciplinary proposals.

% 3. From the experiments and experts evaluations, we show that our model can generate high-quality prediction results and can also reveal the candidate interdisciplines for research with incomplete labels.

% 4. We online a demo of our model, addressed as . In the future, we hope to provide artificial intelligence solutions to the key problem of fund reviewing and allocation based on this framework. 

%Through experiments on HIRPCN, we show that our proposed approach could 
% 4. 在我们的研究能够为基金评审分配这一关键问题提供人工智能的解决方法，这一方法将应用于未来的基金管理工作中并为维护fairness and xxx of xxx 
% We propose a deep learning-based solution for the critical problem of peer-review assignment on the view of detecting the topic path. Our approach will apply in the NSFC funding management in the future and contribute to improving the effectiveness and fairness of the science ecosystem.
% We online a demo of our model for the critical problem of peer-review assignment on the view of detecting the topic path. Our approach will apply in the funding management in the future and contribute to improving the effectiveness and fairness of the science ecosystem.

\begin{itemize}[leftmargin=*]
    \item We identify the critical problems of assigning proposals to its related areas for review. We solve these problems via detecting topic paths on the hierarchical discipline tree. 
    % by using the type-specific semantic information and interdisciplinary knowledge.
    
    % proposal area assignment for review.
    
    % of reviewer-proposal assignment by finding the suitable research domian of the proposal. We solve these problems via detecting topic paths on the hierarchical discipline tree by using the semantic information and interdisciplinary knowledge.
    
    % \item We propose HIRPCN, a framework integrating proposals' semantic information and interdisciplinary knowledge for hierarchical label generation, while can initiatively detect one or more topic paths for interdisciplinary or non-interdisciplinary proposals, resepectively.
    \item We propose HIRPCN, a framework integrating proposals' semantic information and interdisciplinary knowledge for hierarchical label generation, which can generate one or more topic paths for non-interdisciplinary or interdisciplinary proposals, respectively.
    
    % for hierarchical label generation of research proposals using the type-specific semantic information and interdisciplinary knowledge, while can generate multiple topic paths for interdisciplinary proposals.
    
    \item The experiments and experts' evaluations show that our model can generate high-quality prediction results and reveal the candidate interdisciplines for research with incomplete labels.
    
    \item We online a demo\footnote{Demo available in \url{http://101.35.54.56/}} of our model for the critical problem of peer-review assignment on the view of detecting the topic path. Our approach will apply in the funding management in the future and contribute to improving the effectiveness and fairness of the science ecosystem.
\end{itemize}

\section{Definitions and Problem Statement}

\subsection{Important Definitions}

\subsubsection{A Research Proposal}
Applicants  write proposals to apply for grants. 
Figure \ref{Fig.label_1}(1) show a proposal includes multiple documents such as \textit{Title}, \textit{Abstract}, and \textit{Keywords}, \textit{Research Fields}, and more. 
Let's denote a proposal by $A$, the documents in a  proposal are denoted by $D=\{d_t\}_{t=1}^{|T|}$ and the types of each document are denoted by $T =\{t_i\}_{i=1}^{|T|}$. 
The $|T|$ is the total number of the document types, and $d_i$ is the document of $i$-th type $t_i$.  
Every document $d_i$ in the proposal, denoted as $d_i=[w^{1}_i,w^{2}_i,...,w_{i}^{|d_i|}]$, is a sequence of words, where $w^{k}_i$ is the $k$-th word in the document $d_i$. 

\subsubsection{Hierarchical Discipline Structure}\label{d1}
A hierarchical discipline structure, denoted by $\gamma$, is a DAG or a tree that is composed of discipline entities and the directed \textit{Belong-to} relation from a discipline to its sub-disciplines.
% $\mathbf{C}=(C_0, C_1,C_2,...,C_H)$ $C_i=\{c^1_i,c^2_i,...,c^{\tiny |C_i|}_i\}$
The discipline nodes set $\mathbf{C}=\{C_0\cup C_1\cup ...\cup C_H\}$ are organized in $H$ hierarchical levels, where $H$ is the depth of hierarchical level, $C_k=\{c^i_k\}^{|C_k|}_{i=1}$ is the set of the disciplines in the $i$-th level.
%and $|C_i|$ is the total number of disciplines in $C_i$.
The $C_0=\{root\}$ is the root level of $\gamma$. To describe the connection between different disciplines, we introduce $\prec$, a partial order representing the \textit{Belong-to} relationship. $\prec$ is asymmetric, anti-reflexive and transitive\cite{wu2005learning}:
\begin{small}
 \begin{align} 
   & \bullet \text{The only one greatest category }\textit{root}\text{ is the root of the } \gamma,  \nonumber\\
   % root是所有节点的起点
   & \bullet \forall c^x_i \in C_i, c^y_j \in C_j, c^x_i \prec c^y_j \to \space c^y_j \not\prec c^x_i,  \nonumber\\
   % 关系不是对称的
   & \bullet \forall c^x_i \in C_i, c^x_i \not\prec c^x_i, \nonumber\\
   %自身不包含
   & \bullet \forall c^x_i \in C_i, c^y_j \in C_j, c^z_k \in C_k, c^x_i \prec c^y_j \land c^y_j \prec c^z_k \to c^x_i \prec c^z_k.  \nonumber
   % 关系可以传递
   \nonumber
 \end{align}
\end{small} 
Finally, we define the Hierarchical Discipline Structure $\gamma$ as a partial order set $\gamma=(\mathbf{C},\prec)$.
% To describe the connection between different disciplines, we introduce $\prec$, a partial order representing the \textit{Belong To} relationship. 
% $\prec$ is asymmetric, anti-reflexive and transitive\cite{wu2005learning}. 
% The hierarchical discipline structure $\gamma$ over $\mathbf{C}$ can be defined as a partial order set $\gamma=(\mathbf{C},\prec)$.
% \begin{small}
%  \begin{align} 
%   & \bullet \text{The only one greatest category }\textit{root}\text{ is the root of the } \gamma,  \nonumber\\
%   % root是所有节点的起点
%   & \bullet \forall c^x_i \in C_i, c^y_j \in C_j, c^x_i \prec c^y_j \to \space c^y_j \not\prec c^x_i,  \nonumber\\
%   % 关系不是对称的
%   & \bullet \forall c^x_i \in C_i, c^x_i \not\prec c^x_i, \nonumber\\
%   %自身不包含
%   & \bullet \forall c^x_i \in C_i, c^y_j \in C_j, c^z_k \in C_k, c^x_i \prec c^y_j \land c^y_j \prec c^z_k \to c^x_i \prec c^z_k.  \nonumber
%   % 关系可以传递
%   \nonumber
%  \end{align}
% \end{small} 
% Finally, we define the Hierarchical Discipline Structure $\gamma$ over $\mathbf{C}$ as a partial order set $\gamma=(\mathbf{C},\prec)$.
% \end{definition}

% \begin{definition}[\textbf{Co-topic Graph}]
\subsubsection{Interdisciplinary Graph} 
\label{co} 
% \note{modify it, note the interaction between disciplines \& shrink the edges to weight or adj matrix}
A discipline is a combination of the domain knowledge and topics, and the topic of each proposal is carefully selected~\cite{he1999knowledge} and proposed by the scientists and denoted by the \textit{Keywords} text. 
By that, we aim to build a \textit{Interdisciplinary Graph}, denoted as $G=(\mathbf{C}, E)$, to represent the interdisciplinary interactions among disciplines. The $G$ is a collection of discipline nodes $\mathbf{C}$ and directed weighted edges $E=\left\{e_{a\to b}\right\}^{|\mathbf{C}|}_{a,b=1}$. Each $e_{a\to b} \ge 0$ represents the interdisciplinarity from discipline $c_a$ to discipline $c_b$.

So, how to measure \textbf{interdisciplinarity}? S. W. Aboelela~\cite{aboelela2007defining} believe that interdisciplinary research is engaging from two seemingly unrelated fields, which means: (1) these disciplines will have high \textbf{disparity}. (2) the \textbf{frequency} of their related topic will be high.  
Thus, we adopt the \textit{Rao-Stirling}~\cite{stirling2007general} (RS) to measure interdisciplinarity. The RS is a nonparametric quantitative heuristic, which widely adopted to measure the interactions and diversity of the discipline system~\cite{stirling2007general,porter2009science,rafols2010diversity}, ecosystem~\cite{biggs2012toward} and energy security~\cite{winzer2012conceptualizing}, which is defined as:
\begin{equation}
\label{rs}
RS = \sum_{ab(a \not = b)} (p_a\cdot p_b)^\alpha \cdot (d_{ab})^\beta,
\end{equation}
where the first part $\left[\sum_{ab(a \not = b)}(p_a\cdot p_b)\right]$ are proportional representations of elements $a$ and $b$ in the system (frequency), and the rest $\left[\sum_{ab(a \not = b)}d_{a,b}\right]$ is the distance between the $a$ and $b$ (disparity). The $\alpha$ and $\beta$ are two constant to weighting the two components. 
We introduce the RS in a micro view to define the weight on $e_{a\to b}$:
\begin{equation}
\label{wab}
    e_{a \to b} = (p_{a\to b})^\alpha (d_{a\to b})^\beta,
\end{equation}
where $p_{a\to b}$ is the proportional of the proposal in discipline $c_a$ that contains same topics in $c_b$, which represents the penetrating strength of $c_a$ to $c_b$. $d_{a\to b}$ is the topic disparity of $c_a$ in comparing with $c_b$. Their detailed definitions are as:
\begin{equation}
    p_{a\to b} = \frac{\sum_{i}^{n} \mathbbm{1}(k_i \in \{K_a \cap K_b\})F_a[k_i]}{\sum_{i}^{n}F_a[k_i]},
    d_{a\to b} = 1 - \frac{|K_a \cap K_b|}{|K_a|},
\end{equation}
where the $\mathbbm{1}(\cdot) \to \{0, 1\}$ is the indicator function, $K_a$ and $K_b$ are \textit{Keywords}  set of $c_a$ and $c_b$, $n$ is the total number of \textit{Keywords}, and $F_a$ and $F_b$ are the corresponding frequency of the keywords in $K_a$ and $K_b$ appearing in the proposals of their disciplines. In our paper, we assume the two components are equally important and set the $\alpha$ and $\beta$ in Equation~\ref{wab} as 1.

\subsection{Problem Formulation}
% Finally, we provide a formal definition of IRPC task. Given a proposal $A$, we aim to extract its semantic information from the document set $D$ and incorporate the information from the interdisciplinary graph $G$, and assign several label paths for it, where each label path belongs to one of the paths on the hierarchical discipline structure $\gamma$ started from the root node with arbitrary length and each label is the discipline on the paths. Figure 1 demonstrates a toy model of the IRPC problem.

We model the IRPC task in a hierarchical classification schema and use a sequence of discipline-level-specific label sets to represent the proposals' disciplinary codes, denoted as $\mathbb{L} = [L_0, L_1, L_2,...,\\ L_{H_A}]$, where $L_0 = \{l_{root}\}$, $L_i=\{l_i^j\}_{j=1}^{|L_i|}$ is the collection of the labels in the label paths on the $i$-th level, i.e., $\forall l_i^j \in L_i \to l_i^j \in C_i$, $H_A$ is the maximum length of label paths. 
For example, in the Figure~\ref{Fig.label_1}, the ApplyID codes are $F0601$ and $B02$, its labels can be processed as $[\{l_{root}\}, \{F,B\}, \{F06, B02\}, \{F0601\}]$.
To sum up, given the proposal's document set $D$ and the interdisciplinary graph $G$,
we decompose the prediction process into an top-down fashion from the beginning level to a certain level on the hierarchical discipline structure $\gamma$.
Suppose the $k-1$ ancestors labels in $\mathbb{L}$ is $\mathbb{L}_{<k} = [L_{0}, L_1,..., L_{k-1}]$, where $\mathbb{L}_{<1}=\{L_0\}$,
the prediction on level $k$ can be seem as a multi-label classification on $L_k$, formed as:
$$\Omega(D,G,\gamma,\mathbb{L}_{<k};\Theta) \to L_{k}$$ 
where $\Theta$ is the parameters of model $\Omega$.  
Also, to define the probability over the length $H_A$, we add a particular end-of-prediction label $l_{stop}$ into the last set in $\mathbb{L}$, which enables the iterative process of the model to stop on the proper level when the label $l_{stop}$ is predicted. 
Eventually, we can formulate the probability of the assignment of the sequence of label sets for the proposal as:
\begin{equation}
    P(\mathbb{L}|D,G,\gamma;\Theta)=\prod_{k=1}^{H_A} P(L_k|D,G,\gamma,\mathbb{L}_{<k}; \Theta)
    \label{equation_1}
\end{equation}

where $P(\mathbb{L}|D,G,\gamma;\Theta)$ is the overall probability of the proposal $A$ belonging to the label set sequence $\mathbb{L}$, $P(L_k|D,G,\gamma,\mathbb{L}_{<k}; \Theta)$ is the label set assignment probability of $A$ in level-$k$ when given the previous ancestor $\mathbb{L}_{<k}$.
In training, given all the ground truth labels, our goal is to maximize the Equation \ref{equation_1}.

\vspace{-0.3cm}
\section{Interdisciplinary Label Path Detection}

\begin{figure}
\centering
\captionsetup{justification=centering}
\includegraphics[width=0.45\textwidth]{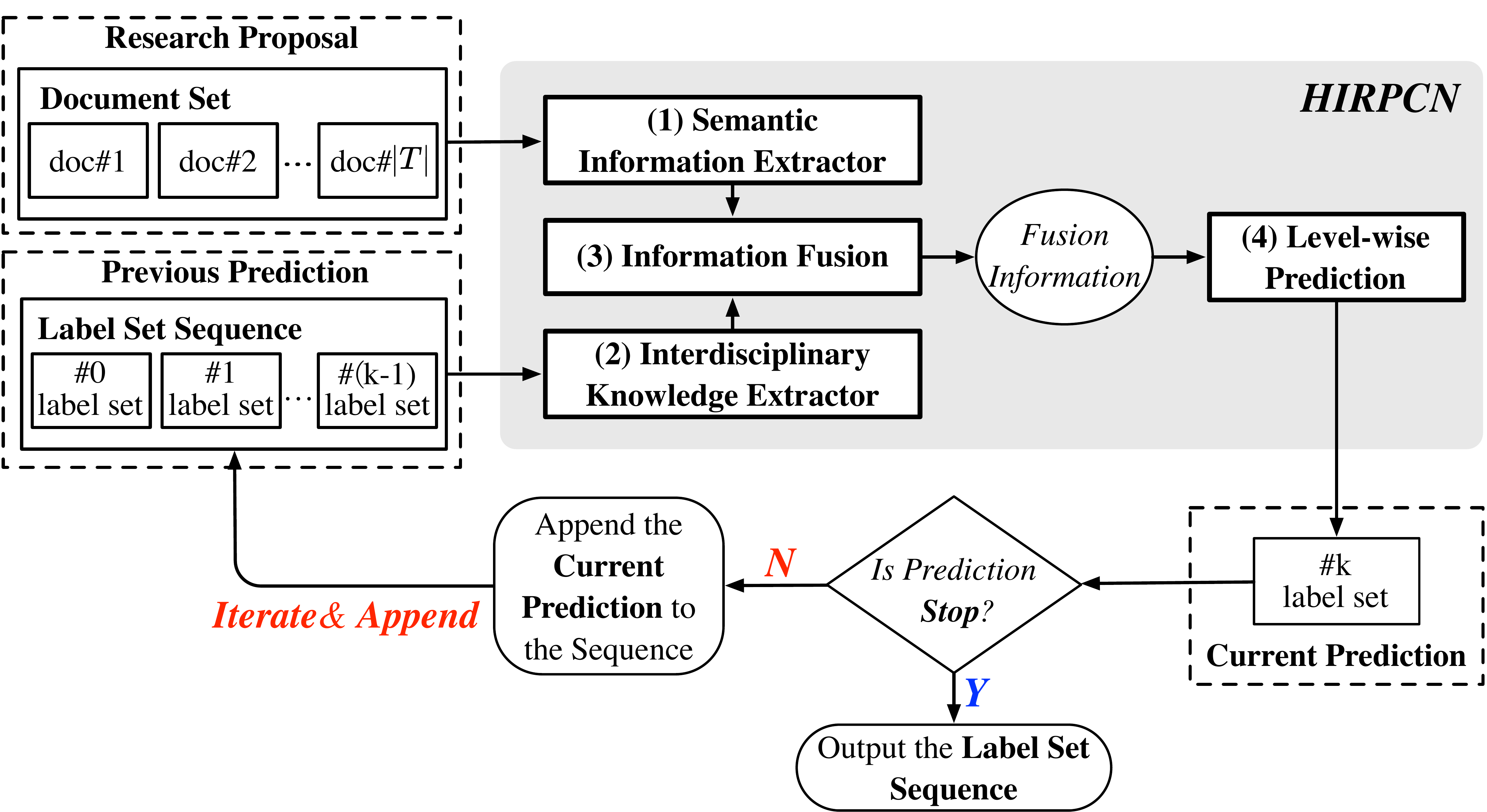} 
\vspace{-0.3cm}
\caption{An overview figure of the HIRPCN during the level-$k$ prediction.}
\label{Fig.model_1}
\vspace{-0.5cm}
\end{figure}

\subsection{Overview of the Proposed Framework} Figure~\ref{Fig.model_1} show our  framework iterate four steps: 
(1) Semantic Information Extractor (SIE), (2) Interdisciplinary Knowledge Extractor (IKE), (3) Information Fusion (IF), and (4) the Level-wise Prediction (LP).
The SIE models the type-specific semantic information (e.g., title, abstract, keywords, research fields) in research proposals.
The IKE learns discipline label embedding at current depth by modeling both interdisciplinary graphs and predicted topics labels of shorter depth. 
The IF fuses the representations of each document in a proposal and topic label embedding.
The LP predicts each discipline's probability to identify the discipline label at the current depth of the three. 
After completing the iteration, our framework takes predicted discipline labels at the current depth and starts to predict fine-grained discipline next depth. 

\begin{figure}[t!h]
\centering
\captionsetup{justification=centering}
\includegraphics[width=0.45\textwidth]{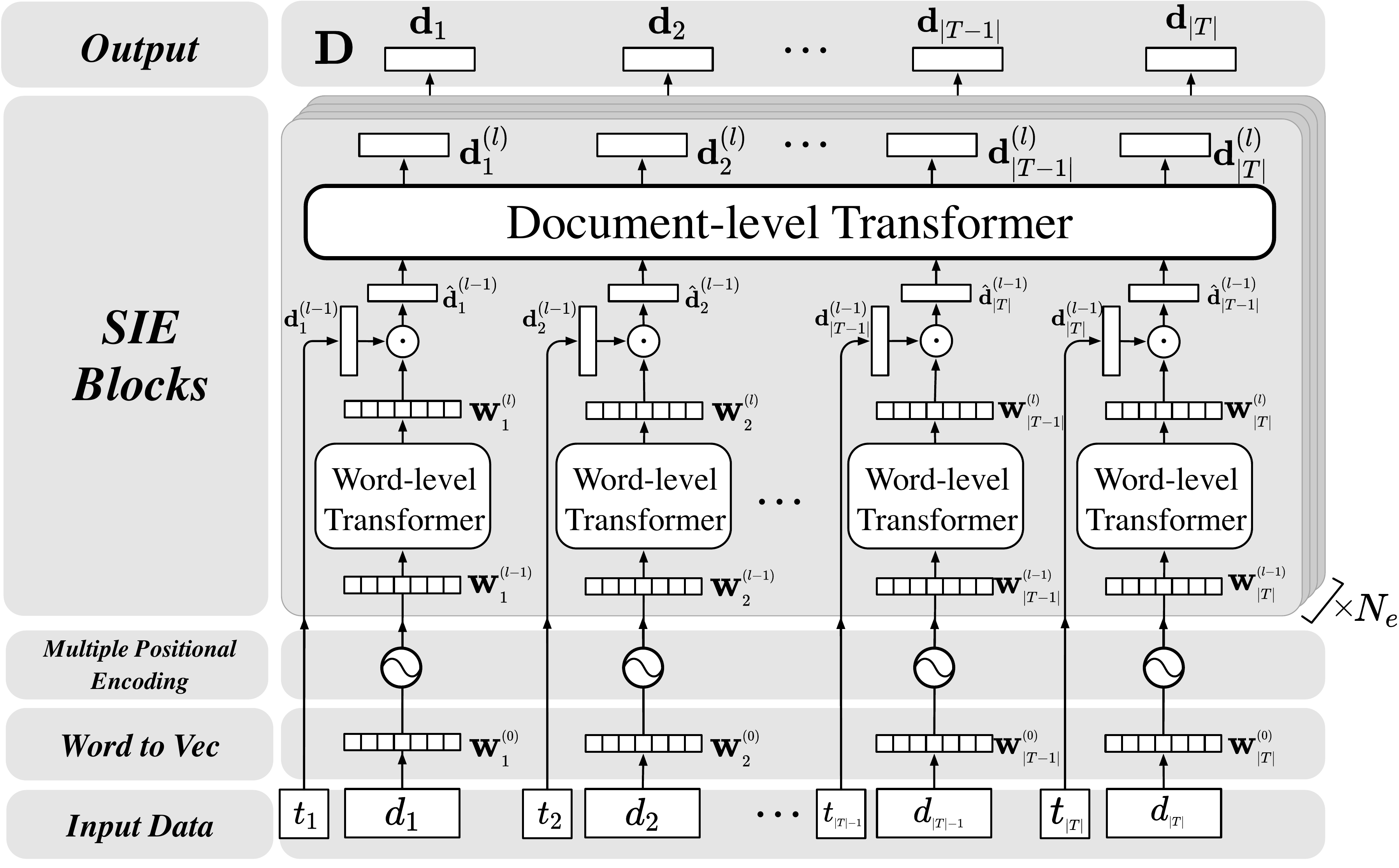}
\vspace{-0.3cm}
\caption{\textit{Semantic Information Extractor}. The $\odot$ means integrating operation between vanilla document representation and their type information.}
\label{Fig.model_encoder}
\vspace{-0.5cm}
\end{figure}

\subsection{Semantic Information Extractor}\label{SIE}
Figure~\ref{Fig.model_encoder} shows an overview of SIE. The basic purpose is to learn the representations of the textual data of a proposal. 
A proposal includes a set of documents (e.g., title, abstract, key words, research fields), each of which includes a sequence of words.
To learn the representation of a proposal, the SIE is structured to include a Word to Vec layer, a Multiple Positional Encoding (MPE) layer, and a SIEBlock that consists of two building blocks:  1) a word-level Transformer~\cite{vaswani2017attention}  and 2) a document-level Transformer. Formally, the embedding of a proposal is learned by:  
\begin{equation}
    \mathbf{D} = N_e \times SIEBlock(\{\mathbf{W}_i^{(0)}\}_{i=1}^{|T|}),
\end{equation}
where the matrix $\mathbf{D}\in \mathbb{R}^{|T| \times h}$ is the embedding of  the textual data and document type information of  the document set of a proposal $D$.  
%$\mathbf{W}^{(0)}=\{\mathbf{w}_1^{(0)}, \mathbf{w}_2^{(0)}, ..., \mathbf{w}_{|T|}^{(0)}\}$ is the initial word embedding set of each documents in $D$. 
$\mathbf{W}_i^{(0)} \in \mathbb{R}^{|d_i| \times h}$ is the word embedding matrix of document $d_i$. The $N_e$ is the total layer number of SIE Block. 
%The $MPE(\cdot)$ means a positional encoding for each word embedding in $\mathbf{W}_i^{(0)}$. 

\noindent\textbf{Our Perspective: Modeling Texts and Document Type for Proposal Embedding.} We consider two essential aspects of a proposal. Firstly, each document (e.g., a title's texts) in a proposal vary in terms of length, and some document is very short. A short concatenated text will cause a severe loss of long-term dependency. 
Secondly, different types (e.g., title, abstract, keywords, research fields) of documents in a research proposal have different weights in identifying the disciplines of the proposal.
Inspired by the long-text modeling methods~\cite{pappagari2019hierarchical,zhang2019hibert,liu2019hierarchical}, we design a \textit{hierarchical transformer architecture}  to model both texts and document type information of a proposal into latent neural embedding. 

\noindent\textbf{Step 1: Word2Vec and Multilple Positional Encoding.} 
% The first input matrix $\mathbf{w}^{(0)}_{i}$ of document $d_i$ is the word embeddings corresponding with $[w^{i}_1,w^{i}_2,...,w^{i}_{|d_i|}]$, 
We convert words of each document in a proposal into initial embeddings (denoted by the matrix $\mathbf{W}^{(0)}_{i}$ of document $d_i$) by pre-trained $h$-dimensional Word2Vec~\cite{mikolov2013distributed} model and sum with positional encoding to preserve the position information.

\noindent\textbf{Step 2: Word-level Transformer.}
In the first stage of SIEBlock, for each document $d_i$, we aim to use a Word-level Transformer to extract the contextual semantic information of the document representation matrix $\mathbf{W}^{(l-1)}_{i}$ of $d_i$ from previous layer to obtain $\mathbf{W}^{(l)}_{i}$, fomulated as:
\begin{equation}
\begin{aligned}
    \mathbf{W}^{(l)}_{i} &= Transformer_{w}(\mathbf{W}^{(l-1)}_i),
\end{aligned}
\end{equation}
where $l$ is the current layer number of SIEBlock, the $\mathbf{W}^{(l)}_i \in \mathbb{R}^{|d_i| \times h}$ is the vanilla document representation matrix and each row is a word embedding. 

% \noindent\textbf{Step 3: Document-level Transformer.}
\noindent\textbf{Step 3: Document-level Transformer.}
In each layer of the Document-level Transformer, we  integrate each vanilla document representation matrix with its correlated type information and feed them into the Transformer together:
\begin{equation}
    \begin{aligned}
        \label{trans2}
        \mathbf{D}^{(l)} = {Transformer}_{d}(\{\mathbf{d}^{(l-1)}_i \odot \mathbf{W}^{(l)}_{i}\}_{i=1}^{|T|}), 
    \end{aligned}
    \end{equation}
where ${Transformer}_{d}(\cdot)$ is the Document-level Transformer, $\mathbf{d}^{(l-1)}_i$ is the type-token vector of $d_i$ in layer-$(l-1)$. In the first layer, each type-token vector $\mathbf{d}^{(0)}_i$ will be initialized randomly. 
Inspired by ViT\cite{dosovitskiy2020image} and TNT\cite{han2021transformer}, we set the $\odot$ operation as: (1) a \textit{Vectorization Operation} on $\mathbf{W}^{(l)}_i$ (2) a \textit{Fully-connected Layer} to transform the vectorized representation from dimension $|d_i| h$ to dimension $h$. (3) an \textit{Element-wise Add} with its type-token vector $\mathbf{d}^{(l-1)}_i$ to get the type-specific representation. $\mathbf{D}^{(l)}\in \mathbb{R}^{h}$ is the outputs of current layer, which can be seen as $|T|$-views of high-level abstraction for proposals.  After $N_e$ times propagation, we can obtain the final output of SIE as $\mathbf{D} = \mathbf{D}^{(N_e)}$.

\begin{figure}[!th]
\centering
\captionsetup{justification=centering}
\includegraphics[width=0.45\textwidth]{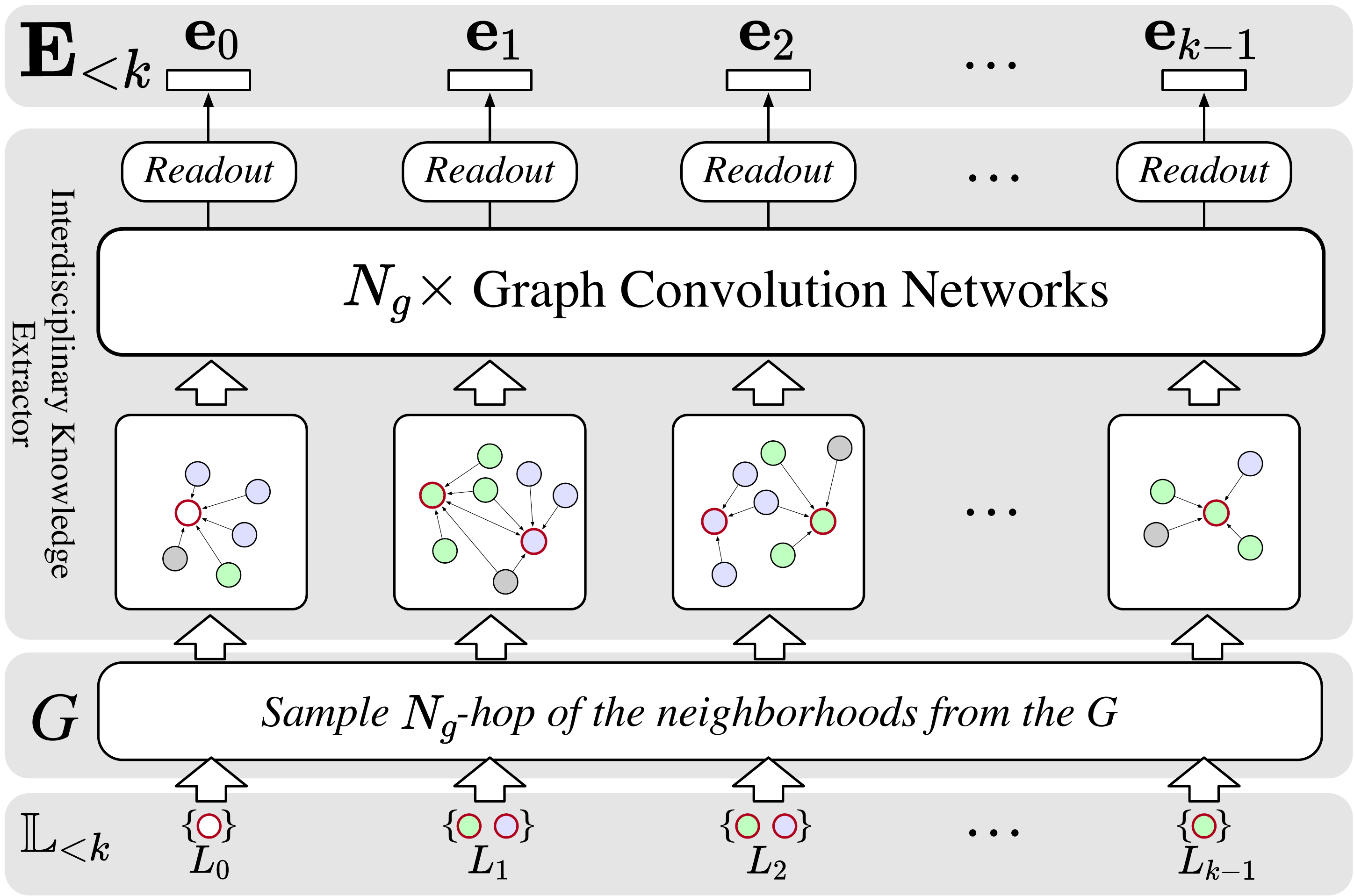} 
\vspace{-0.3cm}
\caption{\textit{Interdisciplinary Knowledge Extractor}. The different colors of the node in $\mathbb{L}_{<k}$ and sub-graph represent different disciplines. The red outline marks each central node.}
\label{Fig.model_encoder_2}
\vspace{-0.6cm}
\end{figure}

\subsection{Interdisciplinary Knowledge Extractor}
% \note{As we aim to infer the Interdisciplinary label, the relations between disciplinary need to be incorporated. The disciplinary tree only contains single disciplinary information, thus, we introduce the co-word graph.}
% Another critical feature is the historical prediction result and its interdisciplinary knowledge.
%在每个迭代预测步中，我们希望利用已经预测出的label信息来帮助下一层的预测，同时我们希望融合交叉学科的信息，因此，我们提出了IKE, 由... 组成，可以被写作...
In each iterative prediction step, we hope to use the previously predicted information (or a given partial label) to help the prediction of the next level. Meanwhile, we hope to form the interdisciplinary knowledge within it. Therefore, we propose IKE to learn previous label embeddings. Figure~\ref{Fig.model_encoder_2} shows an overview of IKE, which mainly consists of multi-layer GCNs and a \textit{Readout Layer}. From an overall perspective, the label embedding is learned by:
\begin{equation}
    \mathbf{E}_{<k} = IKE(\mathbb{L}_{<k}, G),
\end{equation}
% \note{qzy: added $G$}
where the $\mathbf{E}_{<k} \in \mathbb{R}^{k\times h}$ is a label embedding matrix to preserve the interdisciplinary knowledge from previous $k-1$ steps.

\noindent\textbf{Our Perspective: Modeling interdisciplinary interactions for Discipline Embedding.} 
It is conceivable to extract the interactions between disciplines on the Interdisciplinary Graph to acquire the interdisciplinary knowledge, just as model people's behavior by capturing their relation with others on social networks.  % 主谓宾
% In IKE, our fundamental idea is to learn the disciplines embedding from its interdisciplinary interaction with other disciplines.\note{qzy:to discuss. The idea is use graph to model and extract interdisciplinary information} 
To utilize the knowledge, we first model the interdisciplinary relation as $G$, then use Graph Convolutional Networks (GCNs) to aggregate the neighborhood information into the predicted label embedding. 
% 将交叉学科互动融入学科表征
% 我们的观点：Our Perspective: Modeling interdisciplinary interactions for label embeddings. 为了使label embedding获得交叉学科信息。我们首先将交叉关系建模为graph。接着我们使用GNN将有交叉关系的学科聚合到中心学科上，使其获得交叉学科信息。

% Our solution is to represent the predicted label with its sub-graph (to describe its behavior) between each step.

\noindent\textbf{Step 1: Grab the Interactions.} 
Given the sequence of label sets $\mathbb{L}_{<k}$ from previous $k-1$ steps (or from a given partial label from expert), where $\mathbb{L}_{<k} = [L_{0}, L_1,..., L_{k-1}]$. 
Each label set $L_i$ holds the prediction result in the $i$-th level and consists of the discipline labels. 
We first treat each discipline label in every set as a central node and sample its $N_g$-hops sub-graph from the interdisciplinary graph, where the $N_g$ is the layer number of GCNs. Then we feed the sub-graph adjacency weighted matrix and their node features into $N_g$ layers of GCNs:
\begin{equation}
        \mathbf{H}_i = N_g \times GCNLayer(E_i, \mathbf{H}^{(0)}_i),
\end{equation}
where the $\mathbf{H}^{(0)}_i$ is a random initialized feature of nodes on the sub-graph, which is sampled by the $L_i$ as the central nodes, and the $E_i$ is the weighted adjacency matrix of this sub-graph. The $\mathbf{H}_i$ is a hidden output matrix undergoing $N_g$ times GCN layers propagations. 

\noindent\textbf{Step 2: Readout the Features.} 
% \note{QZY: h represents a matrix, should be H. It doesn't feel necessary to divide into two steps.
% If two steps, i suggest step 1 is subgraph sampling and step 2 is subgraph encoding.
% }
% Due to the total number of each prediction result might being varied, we have to pool the matrix and acquire the vectorized feature of each label set. \note{sounds uncomfortable. Then, we readout the label embeddings of $L_i$ from the output matrix:}
Then, we readout the label embeddings of central node set $L_i$ from the output matrix:
% The representation of $L_i$ is obtained by:
\begin{equation}
    \mathbf{e}_i = Readout(\mathbf{H}_i, L_i),
\end{equation}
where the \textit{Readout} layer can be whether a lookup operation or a self-attention layer followed by a mean pooling layer, for simplicity, we set the \textit{Readout Layer} as the former. %a lookup operation followed by a mean pooling to acquire the vectors of $L_i$. 
The $\mathbf{e}_i \in \mathbb{R}^h$ is the final representation of $L_i$. Then we can form the output of IKE as $\mathbf{E}_{<k}=[\mathbf{e}_0, \mathbf{e}_1,...,\mathbf{e}_{k-1}]$. 

The $\mathbf{E}_{<k}$ can be seen as the representation of the prediction results from the previous $k-1$ steps integrated with interdisciplinary knowledge.
We believe that the knowledge helps the model more inclines to predict the discipline that is highly connected to the historical predicted discipline on the interdisciplinary graph, thereby improving the ability to predict the interdisciplines.  
% We believe that the IKE component can help the HIRPCN detect the \textbf{multiple paths} for the IRPC task by utilizing each nodes' neighborhood information. 
% 对交叉学科知识的融合使得模型在下一层预测时更倾向于预测于上层学科联系紧密的学科，从而提高了预测的能力。
% multiple paths --> detect more disciplinary
% label + interdiscplinary knowledge => the knowledge help the proposal with 具有交叉性的先前预测结果，能够在下游预测中预测出与之相关的交叉性的子学科
% \note{qzy: To discuss. Can we go deeper?} 
%GNN 的propagation 使得label embedding 融合了邻域信息，进一步使得模型能够使用上层label的邻域信息做预测，有助于.

\subsection{Information Fusion}
\begin{figure}
\centering
\captionsetup{justification=centering}
\includegraphics[width=0.45\textwidth]{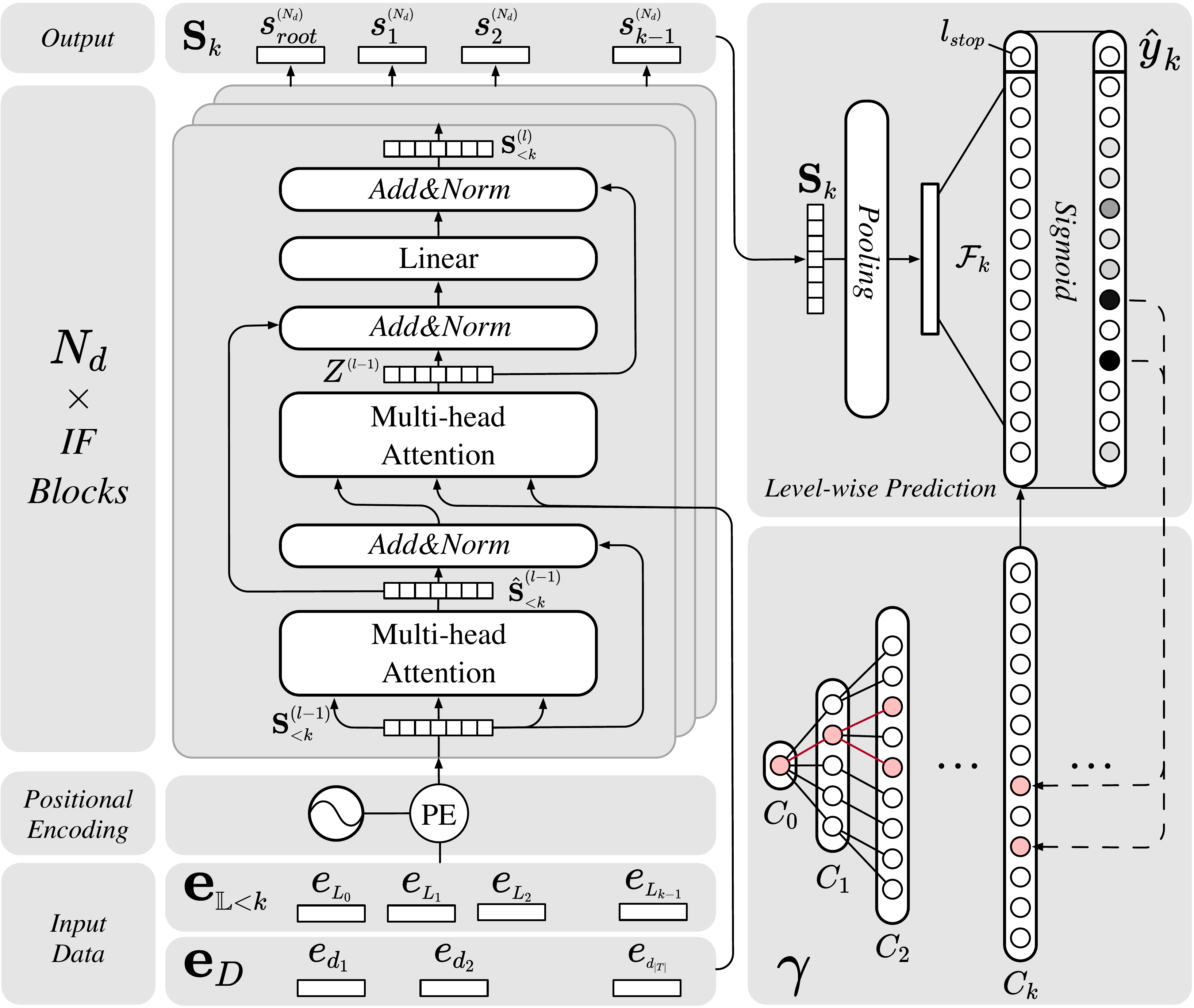}
\vspace{-0.3cm}
\caption{\textit{Information Fusion} and \textit{Level-wise Prediction} in step-$k$. }
% The left side of this figure illustrate the structure of Information Fusion component. The input is the information from previous $k-1$ step prediction and the semantic information of research proposal extracted from SIE. After $N_d$ times iteration of IF Block process, we can obtain the fusion matrix $\mathbf{S}_k$. The right side is the Level-wise prediction component, which will utilize the hierarchical information from hierarchical discipline structure $\gamma$ and the fusion matrix $\mathbf{S}_k$ to generate step-$k$ prediction results.
\label{Fig.model_decoder}
\vspace{-0.6cm}
\end{figure}
In IF, we use a multi-layer \textit{IF Block} to integrate the extracted information from SIE and IKE. As the left side of  Figure~\ref{Fig.model_decoder} shows, the IF consists of a positional encoding and multiple IF Blocks. We start with given its dataflow:
\begin{equation}
    \mathbf{S}_{k} = N_d \times IFBlock(\mathbf{E}_{<k}, \mathbf{D}),
\end{equation}
where $N_d$ is the total layer number of IF Block, and $\mathbf{S}_{k} \in \mathbb{R}^{k\times h}$ is the fusion matrix that integrated the semantic information and the previous prediction information with interdisciplinary knowledge.

\noindent\textbf{Our Perspective: Adaptively Utilize Each Part of Information.} There are two strategies we consider in IF. First, the current prediction state should be significantly affected by its previous results. Second, the model should choose the critical part of semantic information adaptively by its current prediction state. To advance those, we divide the IF into two steps.    

\noindent\textbf{Step 1: Positional Encoding and Previous Label Embedding Fusion.} 
%\note{Incorporate the Label Embedding and Hierarchical Dependency.}
We first sum the $\mathbf{E}_{<k}$ with positional encoding to obtain $\mathbf{S}_{<k}^{(0)}$ for preserving the order of prediction. Then, in layer-{$(l)$} we perform \textit{Multi-Head Self-Attention} 
between each prediction result representation to help each element  in $\mathbf{S}_{<k}^{(l-1)}$ aggregate knowledge from their context: 
\begin{equation}
    \label{M}
    \hat{\mathbf{S}}^{(l)}_{<k} =  \mathbf{S}_{<k}^{(l-1)}\circ \textit{MultiHead}(\mathbf{S}_{<k}^{(l-1)},\mathbf{S}_{<k}^{(l-1)},\mathbf{S}_{<k}^{(l-1)})),
\end{equation}
where  $\circ$ operation is a \textit{Layer Normalization} with a \textit{Residual Connection Layer}.
In this step, the $\textit{MultiHead}(\cdot)$ adaptively aggregate the context information from every step prediction, and the $\circ$ operation integrate the context knowledge to the label embedding and form the the level-$l$ prediction state as $\hat{\mathbf{S}}^{(l)}_{<k} \in \mathbb{R}^{k\times h}$. With the order information, we believe step 1 of IF can capture the hierarchical dependency and the interdisciplinary knowledge in $\mathbb{L}_{<k}$

\noindent\textbf{Step 2: Obtain the Semantic Informaion Adaptively.}
As mentioned before, we hope our model can utilize each part of the research proposal adaptively through prediction progress. Thus, we treat the prediction state $\hat{\mathbf{S}}^{(l)}_{<k}$ as \textit{Query} and the representation of document set  $\mathbf{D}$ as \textit{Key} and \textit{Value} into another \textit{Multi-head Attention} to propagate the semantic information:  
\begin{equation}
    \label{Z}
    \begin{aligned}
        \mathbf{Z}^{(l)} &= \hat{\mathbf{S}}^{(l)}_{<k} \circ \textit{MultiHead}(\hat{\mathbf{S}}^{(l)}_{<k} , \mathbf{D}, \mathbf{D})), \\
        \mathbf{S}_{<k}^{(l)} &= \mathbf{Z}^{(l)} \circ FC(\mathbf{Z}^{(l)}),
    \end{aligned}
\end{equation}
The $\textit{MultiHead}(\cdot)$ adaptively aggregate the semantic information from the research proposal by the current prediction state, and the $\circ$ operation integrate the semantic information to the label embedding and construct the hidden feature $\mathbf{Z}^{(l)} \in \mathbb{R}^{k\times h}$ in level-$l$. After a \textit{Fully-connected Layer} denoted as $FC(\cdot)$ and the $\circ$, we form $\mathbf{S}_{<k}^{(l)}$ to hold the fusion information in level-$l$.  

After $N_d$ times propagations, we acquire the $\mathbf{S}_{k} = \mathbf{S}_{<k}^{(N_d)}$ to represent the output of IF. With the IF, the model will learn the dependency between previous prediction steps and adaptively fuse each part of semantic information by the current prediction progress.
\subsection{Level-wise Prediction}
% According to Equation \ref{equation_1}, we hope that the HIRPCN can generate labels set in a \textbf{top-down fashion} from the beginning level to a \textbf{proper} level $H_A$. 
The right side of Figure \ref{Fig.model_decoder} demonstrate the prediction in step-$k$. In LP, We feed the fusion feature matrix $\mathbf{S}_{k}$ into a \textit{Pooling Layer}, a \textit{Fully-connected Layer}, and a \textit{Sigmoid Layer} to generate each label's probability for $k$-th level-wise label prediction. In our paper, we set this \textit{Pooling Layer} as directly taking the last vector of $\mathbf{S}_{k}$. The formal definition is:
\begin{equation}
 {\hat{\mathbf{y}}}_k= Sigmoid(FC_k(Pooling(\mathbf{S}_{k}))),
\end{equation}
where $\hat{y}^i_k\in {\hat{\mathbf{y}}}_k$ is the predicted probability of the $i$-th discipline $c^i_k \in C_k$ in level-$k$. The $FC_k(\cdot)$ denotes a level-specific feed-forward network with ReLU activation function to project the input to a $|C_k|+1$ length vector. 
After the $Sigmoid(\cdot)$, the final output $\hat{y}_k$ is the probability of $k$-th level's labels. 
Thus, the level-$k$'s objective function $\mathcal{L}_k$ can be defined as:

% \begin{equation}
% \begin{aligned}
%     \mathcal{L}_k(\Theta)    &= \log(P(L_k|D,\gamma,G,\mathbb{L}_{<k}; \Theta))\\
%      &= \sum^{|C_k| + 1}_{i=1} \left(y_i \log(\hat{y}_k^i) + (1 - y_i) \log(1 - \hat{y}_k^i) \right), \\
% \end{aligned}
% \end{equation}
\begin{equation}
    \mathcal{L}_k(\Theta) = \sum^{|C_k| + 1}_{i=1} \left(y^i_k \log(\hat{y}_k^i) + (1 - y^i_k) \log(1 - \hat{y}_k^i) \right),
\end{equation} 
where $y^i_k=\mathbbm{1}(l_k^i \in L_k)$, which aims to discriminate whether the $i$-th label $l_i$ belongs to the truth label set or not. 

The label prediction start from 1st level with given $\mathbb{L}_{<1}$.  
In the $k$-th level prediction, the label corresponding to the index of the value in the $\hat{y}_k$ which achieves the threshold will be selected to construct the current prediction result set $L_k$, and the model will append the $L_k$ to $\mathbb{L}_{<k}$ to construct $\mathbb{L}_{k}$ as current prediction state. If the prediction continues,  $\mathbb{L}_{k}$ will form the next previous ancestor $\mathbb{L}_{<(k+1)}$.
We require that the model can define a distribution over labels of all possible lengths, we add a \textbf{end-of-prediction label} $l_{stop}$ token and set the first element of $\hat{y}_k$ as the probability of $l_{stop}$ in level-k. In the final step, the $l_{stop}$ will include in the prediction result, and the prediction process should end in the level-($H_A$), the model will output $\mathbb{L}_{H_A}$ as the final result. 
%
% Finally, after the prediction from level-$1$ to level-$H_A$, the overall objective function for each proposal $A$ with document set $D$ is to maximize the log-likelihood of the overall probability:
% % \begin{equation}
% %         \Theta = \mathop{\arg\max}_{\Theta}\sum_{k=1}^{H_A} \log P(L_k|D,\gamma,G,\mathbb{L}_{<k}; \Theta)
% % \end{equation}
% \begin{equation}
% \begin{aligned}
%     \mathcal{L}_k(\Theta)    &= \log(P(L_k|D,\gamma,G,\mathbb{L}_{<k}; \Theta))\\
%      &= \sum^{|C_k| + 1}_{i=1} \left(y_i \log(\hat{y}_k^i) + (1 - y_i) \log(1 - \hat{y}_k^i) \right) \\
% \end{aligned}
% \end{equation}
% where the $L_k$ is the ground truth label set of level-$k$ prediction. And $y_i=\mathbbm{1}(l_i \in L_k)$, which aims to discriminate whether the $i$-th label $l_i$ belongs to the truth label set or not.  
\section{Experiments}
In this section, we conduct experiments to evaluate the performance of HIRPCN and answer the following questions: \textbf{Q1.} Is the performance of HIRPCN superior to the existing baseline models? \textbf{Q2.} How is each component of HIRPCN affect the performance? %\textbf{Q3.} Can HIRPCN achieve the best performance to give the finest-grained prediction results? 
\textbf{Q3.} Can HIRPCN achieve the best performance on the prediction at all levels?
\textbf{Q4.} Can the given partial expert advice improve the performance of HIRPCN? \textbf{Q5.} How is the attention mechanism works in each prediction step of HIRPCN? \textbf{Q6.} Can HIRPCN discover the hidden discipline label for interdisciplinary research? 

\vspace{-0.1cm}
\subsection{Experimental Setup}
\subsubsection{Dataset Description}
We collect research proposals written by the scientists from 2020's NSFC research funding application platform, containing 280683  records with 2494 ApplyID code. 
44\% are interdisciplinary research. Among those, 16\% contain two major discipline labels, and 84\% contain one major discipline but show interdisciplinarity in its subdiscipline.
The rest research proposals are marked as non-interdisciplinary research by their applicants. We further organized and divided those proposals by their interdisciplinarity into three datasets named \textit{RP-all}, \textit{RP-bi}, and \textit{RP-differ}. RP-all is the overall dataset with all kinds of research proposals. RP-bi consists of the research proposal whether it exhibits interdisciplinarity in the subdiscipline or their major discipline. RP-differ only includes the research proposal with two major disciplines.  
\subsubsection{Evaluation Metrics}
To fairly measure the HIRPCN with other baselines, we evaluate the prediction results with several widely adopted metrics \cite{metric1,metric2,xiao2021expert} in the domain of Multi-label Classification, i.e., the \textit{Micro-F1} (MiF1), and \textit{Macro-F1} (MaF1).
\subsubsection{Baseline Algorithms}
We compare our model with seven Text Classification (TC) methods including TextCNN~\cite{textcnn}, DPCNN~\cite{dpcnn},  FastText~\cite{fasttext1,fasttext2},  TextRNN~\cite{textrnn},  TextRNN-Attn~\cite{textrnnattn},  TextRCNN~\cite{textrcnn},  Trasnformer~\cite{vaswani2017attention} and three state-of-the-art Hierarchical Multi-label Classification (HMC) approaches including  HMCN-F,  HMCN-R~\cite{wehrmann2018hierarchical},  HARNN~\cite{huang2019hierarchical}. 
We also post three ablation models of HIRPCN as baselines, including  w/o Interdisciplinary Graph (w/o IG),  w/o Hierarchical Transformers (w/o HT), and  w/o All. 
% The HMC methods, HIRPCN, and the ablation models are hierarchy-aware and follow the top-down fashion. HIRPCN and w/o IG are text-type aware. HIRPCN and w/o HT are equipped with the Interdisciplinary Graph and are mindful of interdisciplinary knowledge. w/o ALL replaces the SIE with a vanilla Transformer and removes the IKE component, and we use it to demonstrate the architecture performance.\note{discuss the ablation setting. A replace .. with .., B remove and .. use .., C replace both.} 
For all the experiments, we conduct 5-fold cross-validation and report the average recommendation performance. 
\subsubsection{Hyperparameters, Source Code and Reproducibility}
We set the SIE layer number $N_e$ to 8, the dimension size $h$ to 64, and the multi-head number to 8. The IKE layer number $N_g$ is set to 1. The IF layer number $N_d$ is set to 8, and the multi-head number is set to 8. We use Word2Vec\cite{mikolov2013distributed} model with a dimension ($h$) 64 to generate the word embedding for each Chinese characters. For the detail of training, we use Adam optimizer\cite{kingma2014adam} with a learning rate of $1\times 10^{-3}$, and set the mini-batch as 512, adam weight decay as $1 \times 10^{-7}$. The dropout rate is set to $0.2$ to prevent overfitting. The warm-up step is set as 1000.
We have shared the source code via Dropbox\footnote{\url{https://www.dropbox.com/sh/x5m1jlcax8jp6tk/AAAg-KrGM8cHZuqoSfHKLRC0a}}.
\subsubsection{Environmental Settings}
All methods are implemented by PyTorch 1.8.1\cite{paszke2019pytorch}. The experiments are conducted on a CentOS 7.1 server with a AMD EPYC 7742 CPU and 8 NVIDIA A100 GPUs.

\vspace{-0.1cm}
\subsection{Experimental Results}
\subsubsection{\textbf{RQ1:} Overall Comparison}
\begin{table}
\centering
\captionsetup{justification=centering}
\caption{Experimental results on all dataset. The best results are highlighted in \textbf{bold}. The second-best results are highlighted in \underline{underline}.}
\label{overall}

\resizebox{0.45\textwidth}{!}{%
\begin{tabular}{l|cc|cc|cc}
\hline
\multicolumn{1}{l|}{Dataset}                                        & \multicolumn{2}{c|}{RP-all}                                              & \multicolumn{2}{c|}{RP-bi}                                               & \multicolumn{2}{c}{RP-differ}                                           \\ \hline
\multicolumn{1}{l|}{Method}   &\multicolumn{1}{l|}{MiF1}            & \multicolumn{1}{l|}{MaF1} & \multicolumn{1}{l|}{MiF1}            & \multicolumn{1}{l|}{MaF1} & \multicolumn{1}{l|}{MiF1}            & \multicolumn{1}{l}{MaF1} \\ \hline
\multicolumn{1}{l|}{TextCNN}                 & \multicolumn{1}{c|}{0.456}               & 0.168                         & \multicolumn{1}{c|}{0.426}               & 0.148                         & \multicolumn{1}{c|}{0.438}               & 0.063                        \\
\multicolumn{1}{l|}{DPCNN}                            & \multicolumn{1}{c|}{0.376}               & 0.108                         & \multicolumn{1}{c|}{0.364}               & 0.080                         & \multicolumn{1}{c|}{0.346}               & 0.020                        \\
\multicolumn{1}{l|}{FastText}                            & \multicolumn{1}{c|}{0.460}               & 0.167                         & \multicolumn{1}{c|}{0.427}               & 0.149                         & \multicolumn{1}{c|}{0.432}               & 0.067                        \\

\multicolumn{1}{l|}{TextRNN}                                & \multicolumn{1}{c|}{0.402}               & 0.100                         & \multicolumn{1}{c|}{0.364}               & 0.077                         & \multicolumn{1}{c|}{0.351}               & 0.021                        \\

\multicolumn{1}{l|}{TextRNN-Attn}                & \multicolumn{1}{c|}{0.413}               & 0.098                         & \multicolumn{1}{c|}{0.367}               & 0.072                         & \multicolumn{1}{c|}{0.338}               & 0.018                        \\
\multicolumn{1}{l|}{TextRCNN}                             & \multicolumn{1}{c|}{0.426}               & 0.138                         & \multicolumn{1}{c|}{0.381}               & 0.100                         & \multicolumn{1}{c|}{0.367}               & 0.025                        \\
\multicolumn{1}{l|}{Transformer}                           & \multicolumn{1}{c|}{0.370}               & 0.083                         & \multicolumn{1}{c|}{0.339}               & 0.066                         & \multicolumn{1}{c|}{0.358}               & 0.023                        \\ \hline
\multicolumn{1}{l|}{$\text{HMCN-F}$}                                 & \multicolumn{1}{c|}{0.676}               & 0.459           & \multicolumn{1}{c|}{0.582} & 0.308           & \multicolumn{1}{c|}{0.569}               & $\underline{0.161}$          \\
\multicolumn{1}{l|}{$\text{HMCN-R}$}                             & \multicolumn{1}{c|}{0.599}               & 0.289                         & \multicolumn{1}{c|}{0.506}               & 0.185                         & \multicolumn{1}{c|}{0.478}               & 0.069                        \\
\multicolumn{1}{l|}{HARNN}                                          & \multicolumn{1}{c|}{0.686} & 0.443                         & \multicolumn{1}{c|}{0.574}               & 0.294                         & \multicolumn{1}{c|}{0.516}               & 0.102                        \\ \hline
% \multicolumn{1}{l|}{$\text{HIRPCN w/o ALL}$}                         & \multicolumn{1}{c|}{0.436}               & 0.168                         & \multicolumn{1}{c|}{0.435}               & 0.164                         & \multicolumn{1}{c|}{0.485}               & 0.089                        \\
\multicolumn{1}{l|}{$\text{w/o ALL}$}                         & \multicolumn{1}{c|}{0.704}               & 0.429                        
& \multicolumn{1}{c|}{0.583}               & 0.259 % epoch 270                         
& \multicolumn{1}{c|}{0.512}               & 0.098 % epoch 820 
\\
\multicolumn{1}{l|}{$\text{w/o IG}$}                              & \multicolumn{1}{c|}{\underline{0.737}}               & \underline{0.488}                         & \multicolumn{1}{c|}{\underline{0.636}}               & \underline{0.329}                         & \multicolumn{1}{c|}{$\underline{0.577}$} & 0.127                        \\
\multicolumn{1}{l|}{$\text{w/o HT}$}                              & \multicolumn{1}{c|}{0.710}               &  0.438                         & \multicolumn{1}{c|}{0.587}                &  0.276                        & \multicolumn{1}{c|}{0.533} & 0.099   
\\ \hline
\multicolumn{1}{l|}{$\text{HIRPCN}$}                            & \multicolumn{1}{c|}{$\textbf{0.748}$}    & $\textbf{0.512}$              & \multicolumn{1}{c|}{$\textbf{0.648}$}    & $\textbf{0.384}$              & \multicolumn{1}{c|}{$\textbf{0.635}$}    & $\textbf{0.180}$             \\ \hline
\end{tabular}}%
% differ
% ntng
% EP_test:2890: 100%|| 8/8 [00:01<00:00,  7.10it/s]
% [Local] epoch 2890 Predict by threshold in Level-0 - Micro: F1 0.81044
% [Local] epoch 2890 Predict by threshold in Level-0 - Macro: F1 0.692982
% [Local] epoch 2890 Predict by threshold in Level-1 - Micro: F1 0.564913
% [Local] epoch 2890 Predict by threshold in Level-1 - Macro: F1 0.441507
% [Local] epoch 2890 Predict by threshold in Level-2 - Micro: F1 0.357275
% [Local] epoch 2890 Predict by threshold in Level-2 - Macro: F1 0.133318
% [Local] epoch 2890 Predict by threshold in Level-3 - Micro: F1 0.219862
% [Local] epoch 2890 Predict by threshold in Level-3 - Macro: F1 0.0606127
% [Local] epoch 2890 Predict by threshold in overall - Micro: F1 0.526637
% [Local] epoch 2890 Predict by threshold in overall - Macro: F1 0.10956

% ntwg
% EP_test:2550: 100%|| 8/8 [00:01<00:00,  6.31it/s]
% [Local] epoch 2550 Predict by threshold in Level-0 - Micro: F1 0.812279
% [Local] epoch 2550 Predict by threshold in Level-0 - Macro: F1 0.703991
% [Local] epoch 2550 Predict by threshold in Level-1 - Micro: F1 0.551742
% [Local] epoch 2550 Predict by threshold in Level-1 - Macro: F1 0.445986
% [Local] epoch 2550 Predict by threshold in Level-2 - Micro: F1 0.369235
% [Local] epoch 2550 Predict by threshold in Level-2 - Macro: F1 0.138233
% [Local] epoch 2550 Predict by threshold in Level-3 - Micro: F1 0.223552
% [Local] epoch 2550 Predict by threshold in Level-3 - Macro: F1 0.0650199
% [Local] epoch 2550 Predict by threshold in overall - Micro: F1 0.528945
% [Local] epoch 2550 Predict by threshold in overall - Macro: F1 0.114216
\vspace{-0.3cm}
\end{table}
We first evaluate all methods on all datasets in Table~\ref{overall}. In this comparison, we organized every level-wise prediction result flatly, then we used the \textit{MiF1} and the \textit{MaF1} as the evaluation metrics. From the results, we can observe that:
% \note{++some discusion about other select methods}

(1) Our proposed method, HIRPCN achieves the best performance on all datasets in all evaluation metrics, proving that HIRPCN can better solve the challenges mentioned in the IRPC task.

% (2) The HMC methods perform better than TC methods. The reason is that the label set is numerous in a flat view than a hierarchical view.\note{why?  a flat view can not learn dependency.} Further, considering the hierarchy in the discipline system is indispensable, proving that the HMC schema is suitable for the IRPCN task. \note{HMT, dependency HMC better than TC}
(2) The HMC methods perform better than TC methods. The reason is that the discipline system exhibits a hierarchical structure. HMC methods organize the labels in a hierarchical rather than a flat view, better capturing the hierarchical dependency between labels than TC methods.

% (3) It is worth noting that, from RP-all to RP-differ, all the methods' performance decrease. The reason is that the interdisciplinarity among each dataset and the label number on each level increase which causes the classification problem to become hard to solve. \note{wired, make the problem harder.}
(3) It is worth noting that the performance of all the methods decreases on RP-bi and RP-differ compared with that on RP-all. The reason is that interdisciplinary proposals have more labels on each level, making the classification task harder.
The performance on RP-bi is generally better than that on RP-differ, indicating that predicting two major disciplines is much more challenging as the disparity on their domain.

\subsubsection{\textbf{RQ2:} Ablation Study}
We also discuss the performance of each ablation variant of HIRPCN.  The w/o IG removes the IKE component, so in each prediction step, w/o IG will have no awareness of the interdisciplinary knowledge. The w/o HT replaces the SIE component with a vanilla Transformer, which makes the model unable to discriminate the different types of documents and concatenate every word token sequence together. The w/o ALL ablate the IKE and SIE to represent the architecture's performance. From the Figure~\ref{overall}, we can observe that:
(1) The results of HIRPCN and its three variant models on three datasets prove that both the SIE and IKE components can improve the model's performance. w/o HT achieves worse results than w/o IG, indicating that SIE contributes more to the task than IKE.

(2) w/o IG performances slightly worse (e.g., -1.5\% deterioration on Micro-F1) than HIRPCN on the RP-all dataset, while the margin (e.g., -1.9\% \& -9.1\% deterioration on Micro-F1) becomes larger on RP-bi and RP-differ. This phenomenon proves that the \textbf{interdisciplinary interaction topology information} plays a significant role in the classification of interdisciplinary proposals.

(3) HIRPCN and w/o IG, both with hierarchical Transformer, are significant superior to other HMC methods on RP-all and RP-bi, showing the advantage of modeling \textbf{various lengths of the documents} and the \textbf{multiple types of text} for proposal embeddings. On RP-differ, w/o IG performs slightly worse than HMCN-F due to the replacement of IKE. 
% }
% 

% 有的标签在数据集里根本不存在 =》 在这个标签上的分数是0 =》 average之后的值很低
% (5) We observe that all methods perform worse at Macro-F1 than Micro-F1. We carefully check the datasets and the main reason is that not all disciplines have an interdisciplinary interaction with others. Thus, the label set of RP-differ and RP-bi might not be cover all labels, which causes the Macro-F1 score lower. \note{this discussion may be removed}
\subsubsection{\textbf{RQ3:} Level-wise Performance}
\begin{table}
\centering
\captionsetup{justification=centering}
\caption{Experimental results on RP-differ in level-wise. The best results are highlighted in \textbf{bold}. The second-best results are highlighted in \underline{underline}.}
\vspace{-0.1cm}
\label{RPdiffer}
\resizebox{0.45\textwidth}{!}{%
\begin{tabular}{l|cc|cc|cc|cc}
\hline
\multicolumn{1}{l|}{\multirow{2}{*}{Method}} & \multicolumn{2}{c|}{Level-1}                                  & \multicolumn{2}{c|}{Level-2}                                  & \multicolumn{2}{c|}{Level-3}                                  & \multicolumn{2}{c}{Level-4}                   \\ \cline{2-9} 
\multicolumn{1}{c|}{}                        & \multicolumn{1}{c|}{MiF1} & \multicolumn{1}{c|}{MaF1} & \multicolumn{1}{c|}{MiF1} & \multicolumn{1}{c|}{MaF1} & \multicolumn{1}{c|}{MiF1} & \multicolumn{1}{c|}{MaF1} & \multicolumn{1}{c|}{MiF1} & \multicolumn{1}{c}{MaF1}\\ \hline
TextCNN                 & 0.735          & 0.600            & 0.403          & 0.285          & 0.254          & 0.067          & 0.190           & 0.040          \\
DPCNN                   & 0.689          & 0.543          & 0.284          & 0.151          & 0.132          & 0.017          & 0.084          & 0.009         \\
FastText                & 0.730           & 0.610           & 0.402          & 0.303          & 0.251          & 0.078          & 0.164          & 0.037         \\
TextRNN                 & 0.692          & 0.569          & 0.296          & 0.155          & 0.130           & 0.018          & 0.073          & 0.009         \\
TextRNN-Attn            & 0.677          & 0.548          & 0.278          & 0.148          & 0.125          & 0.015          & 0.071          & 0.008         \\
TextRCNN                & 0.690           & 0.559          & 0.310           & 0.183          & 0.165          & 0.025          & 0.105          & 0.010          \\
Transformer             & 0.702          & 0.577          & 0.308          & 0.183          & 0.135          & 0.022          & 0.070           & 0.009         \\ \hline
HMCN-F                    & 0.782          & 0.670           & 0.565          & 0.461          & \underline{0.460}     & \underline{0.201}    & \underline{0.341}    & \underline{0.102}   \\
HMCN-R                  & 0.770           & 0.664          & 0.478          & 0.339          & 0.278          & 0.074          & 0.183          & 0.041         \\
HARNN                   & 0.793          & 0.683          & 0.529          & 0.409          & 0.341          & 0.122          & 0.226          & 0.058         \\ \hline
% HMDG w/o ALL                 & 0.797          & 0.684          & 0.524          & 0.410           & 0.321          & 0.107          & 0.206          & 0.045         \\
w/o ALL                 & 0.810          & 0.692          & 0.564          & 0.441           & 0.357          & 0.133          & 0.219          & 0.060         \\
w/o IG                 & \underline{0.825}    & \underline{0.715}    & \underline{0.627}    & \underline{0.506}    & 0.429          & 0.150           & 0.281          & 0.075         \\
w/o HT                   & 0.812 & 0.703 & 0.551 & 0.445 & 0.369 & 0.138 & 0.223 & 0.065\\
\hline
HIRPCN                    & \textbf{0.846} & \textbf{0.731} & \textbf{0.658} & \textbf{0.540} & \textbf{0.519} & \textbf{0.222} & \textbf{0.368} & \textbf{0.114}\\
\hline
\end{tabular}}%
\vspace{-0.2cm}
\end{table}
% \note{to discuss why the level results can reflect the granularity.}
%We further evaluate the ability of HIRPCN and other baselines to give the proper granularity of prediction. We use the level-wise MiF1 and level-wise MaF1 as metrics to indicate the performance on granularity of the prediction. We compare HIRPCN with all baselines on RP-differ. Figure~\ref{RPdiffer} shows that HIRPCN outperforms other baseline methods and variants on each level. We also observe that the accuracy of each model tends to decrease, which is caused by the number of categories increasing rapidly when the hierarchical level depth increases. The results on every level prove that HIRPCN could predict the best \textbf{granular} on IRPC task.
% 1/ level 结果都好
% 2/ 探讨下降原因HMC方法
% 3/ 下降幅度更低

%我们进一步展示了模型和baseline细粒度的分类结果 by report他们每一层的分类结果。图2展示了结果。我们发现每个level结果都好，证明我们的模型

We further report the performance of level-wise prediction of HIRPCN and other baselines on the RP-differ dataset. 
Figure 2 shows the level-wise Micro-F1 and level-wise Macro-F1 of classification results on four levels.
From the results, we can observe that  HIRPCN outperforms all the baseline methods and the variants on every level, showing the advanced ability on the classification at all granularity.
We also observe that the performance of each model tends to decrease with the depth of level increase because the number of categories increases rapidly, resulting in the classification task becoming harder.
Lastly, w/o IG demonstrates a relatively slow decreasing tendency compared with other variants, proving the advantage of incorporating interdisciplinary knowledge for fine-grained discipline prediction.

% \subsection{Case Study}
% We want to thoroughly analyze the model and discuss the potential for applying an online system. During the case study, we introduce three unique model features which can significantly improve the ideology of the funding application: 1) Hidden Interdiscipline Find 2) Prediction with Expert Knowledge 3) The Granularity. Suppose we are a newbie to apply for NSFC funding, and we might have no idea about the whole complex discipline system or the ApplyID code. The HIRPCN will provide the recommended discipline and help scientists choose the ApplyID code level by level with proper granularity. On the funding administrator side, they hope the HIRPCN could automatically mark the interdisciplinary proposals and the related disciplines to further decide on reviewer assignment.

\subsubsection{\textbf{RQ4:} Prediction with Given Labels}
\begin{figure}
\centering
\captionsetup{justification=centering}
\includegraphics[width=0.45\textwidth]{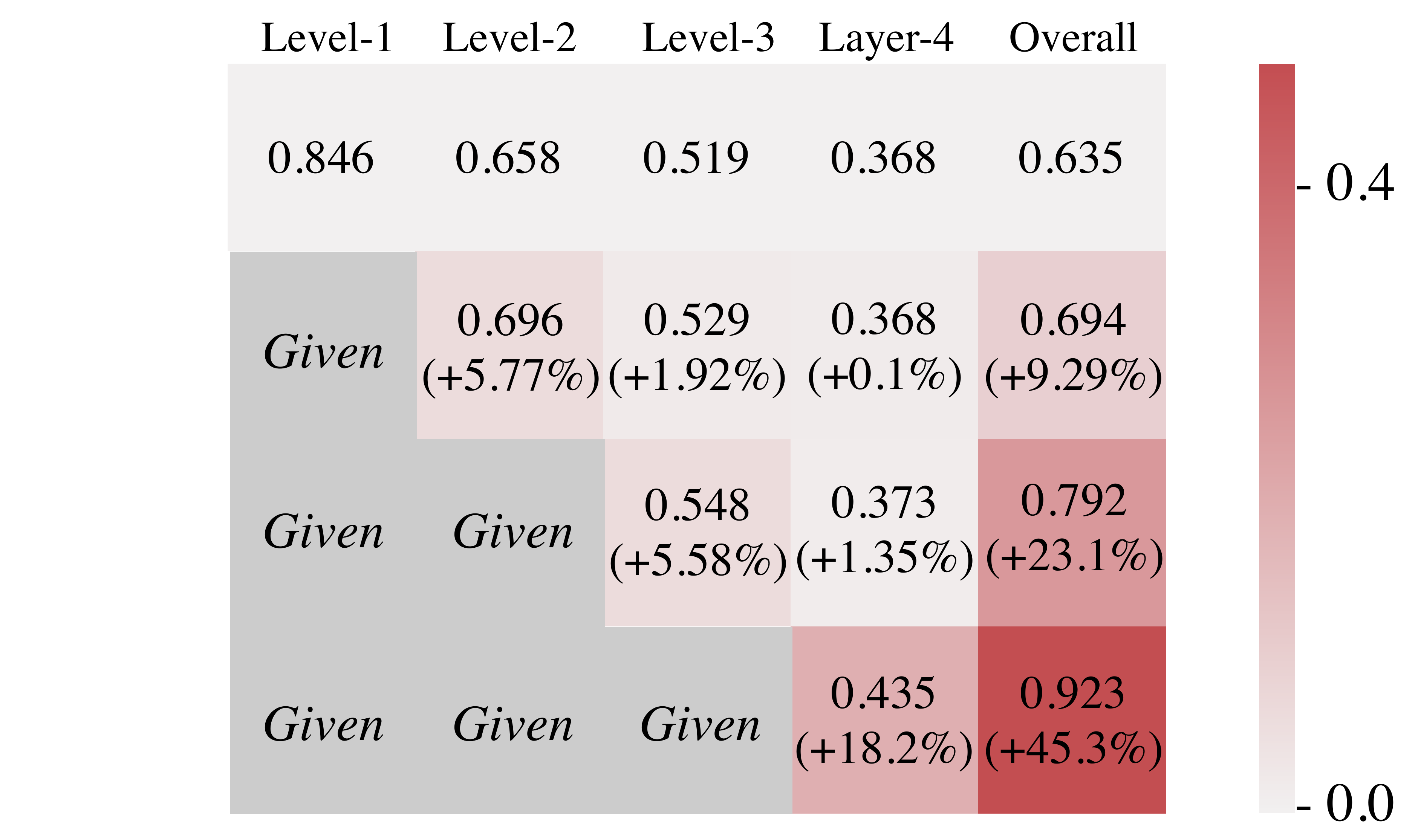}
\vspace{-0.3cm}
\caption{The performance with given labels.}
\label{Fig.partial_label}
\vspace{-0.3cm}
\end{figure}

The \textit{Given Labels} are a partial label set provided by scientists or funding administrators when they try to fill the ApplyID. As we mentioned in the previous section, we hope the HIRPCN can be an assistant who knows the whole picture of the discipline system. This case study evaluates the improvement when HIRPCN receives the incomplete labels at each level and predicts the rest label. It is also worth noting that other baseline methods use a hidden vector to represent previous prediction results, which cannot initialize the prediction from any given partial label set. Thus, we compared the performance of HIRPCN with different levels of provided expert knowledge on the RP-differ dataset. We evaluate the model by the level-wise MiF1 and the overall MiF1. As you can see in Figure~\ref{Fig.partial_label}, the first row of the figure shows the performance of the HIRPCN on RP-differ when no expert knowledge is given. The rest rows illustrate when the first level, first and second level, first, second and third level are given, how the model improves its performance by using these partial labels. The color of each cell shows the improvement when expert knowledge is given. From the heatmap, we can observe that:

(1) With the incomplete labels given, the overall performance of HIRPCN raised. This phenomenon shows that the architecture guarantees the model can utilize the partial label and can improve the overall performance.

(2) We notice that the given labels can significantly improve the next-level prediction but declines in the remaining level. This phenomenon is reasonable because the model will pay more attention to the previous level than other earlier-level (as the discussion of attention in Appendix \ref{appendix:attention} shows). 

\subsubsection{\textbf{RQ5:} Attention Mechanism Explain} 
\begin{figure}
\centering
\captionsetup{justification=centering}
\includegraphics[width=0.45\textwidth]{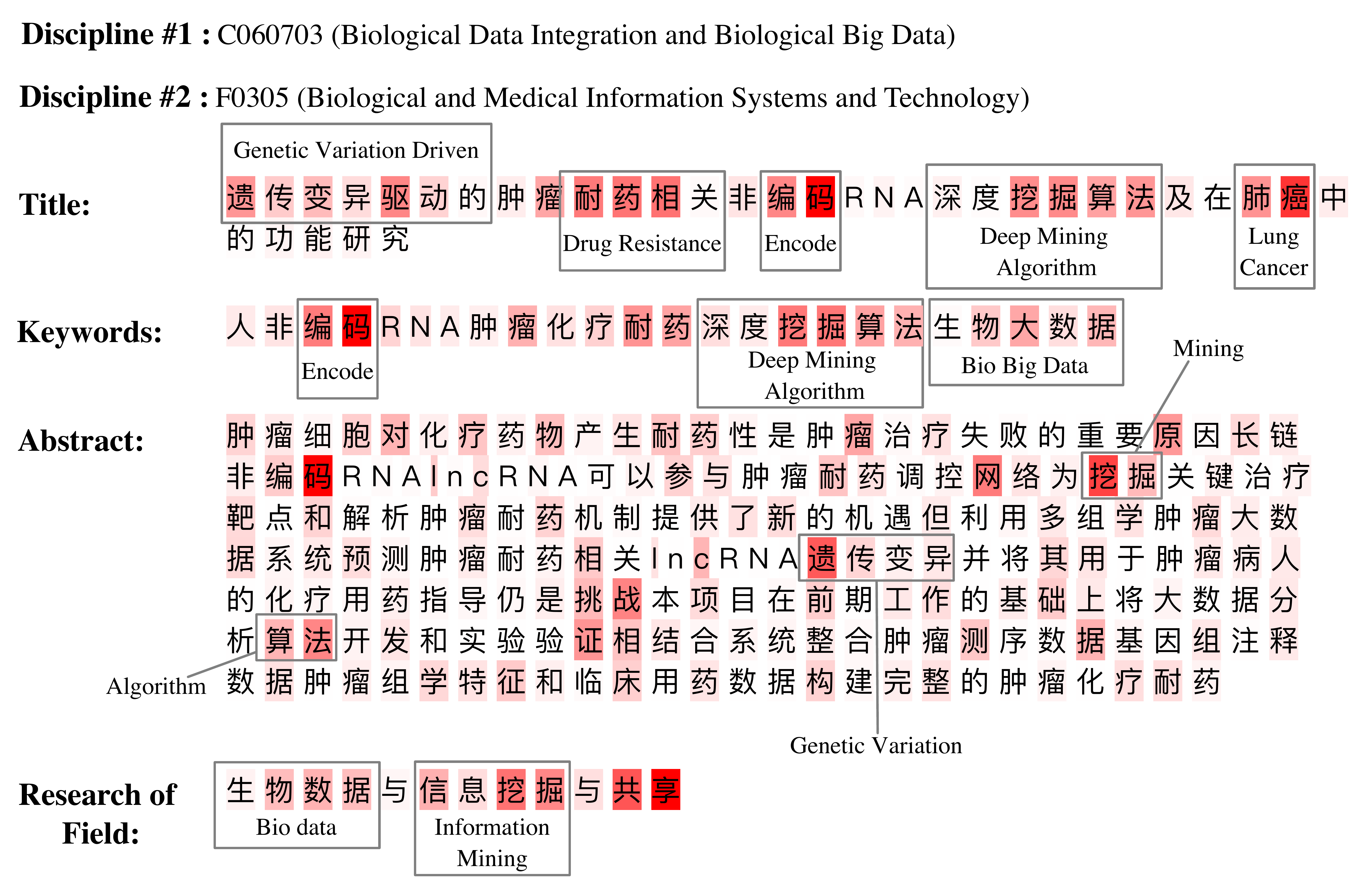} 
\vspace{-0.3cm}
\caption{A interdisciplinary research proposal and its word-level attention visualization from the \textit{Information Sciences} and the \textit{Life Sciences}.}
\label{Fig.attn_text}
\vspace{-0.5cm}
\end{figure}
We introduce the attention mechanism in most parts of HIRPCN. In this section, we visualize the attention value learned by the first layer of Word-Level Transformer (further discussion of other attention can be found at Appendix~\ref{appendix:attention}). Figure~\ref{Fig.attn_text} is a case study of an interdisciplinary research proposal. 
As we mentioned in Section~\ref{SIE}, each representation of the word token will be aggregated by different attention values. If a word token gains more SIE's attention, they will be shading redder in this figure. From the figure, we can observe some interesting phenomena even we take word token as the smallest unit in our model:

(1) As we hope, in the SIE, the model will pay more attention to the critical word but not the stop words, especially in the long text data. For example, in the \textit{Abstract}, the word “挖掘” (Mining) and “算法” (Algorithm) will generally gain more attention than the word “是” (Is) and “和” (And). This explains that the HIRPCN can discriminate the importance of each word token initiatively and find the keywords. 

(2) Further, the attention value will be high for the word token relevant to the \textbf{corresponded discipline} of this research proposal. For example, in the \textit{Title}, the word “肺癌” (Lung Cancer) and “深度挖掘算法” (Deep Mining Algorithm) will gain more attention due to its relation to the \textit{Life Sciences} and \textit{Information Sciences}, respectively. We believe this mechanism helps the model to make predictions depending on the semantic information from those critical words. This phenomenon shows that the HIRPCN will extract the useful semantic information from documents by highlight domain-related words.  

(3) The self-attention mechanism works well in the text with multiple types and variable-length, whether the “耐药相关” (Drug Resistance) in the \textit{Title} or the “挖掘” (Mining), “算法” (Algorithm) in the \textit{Abstract}. We believe this phenomenon proves that the architecture of SIE is superior for handling the complex structure in research proposals. 
% \vspace{-0.1cm}
\subsubsection{\textbf{RQ6:} Hidden Interdiscipline Find} 
\begin{figure}
\centering
\captionsetup{justification=centering}
\includegraphics[width=0.45\textwidth]{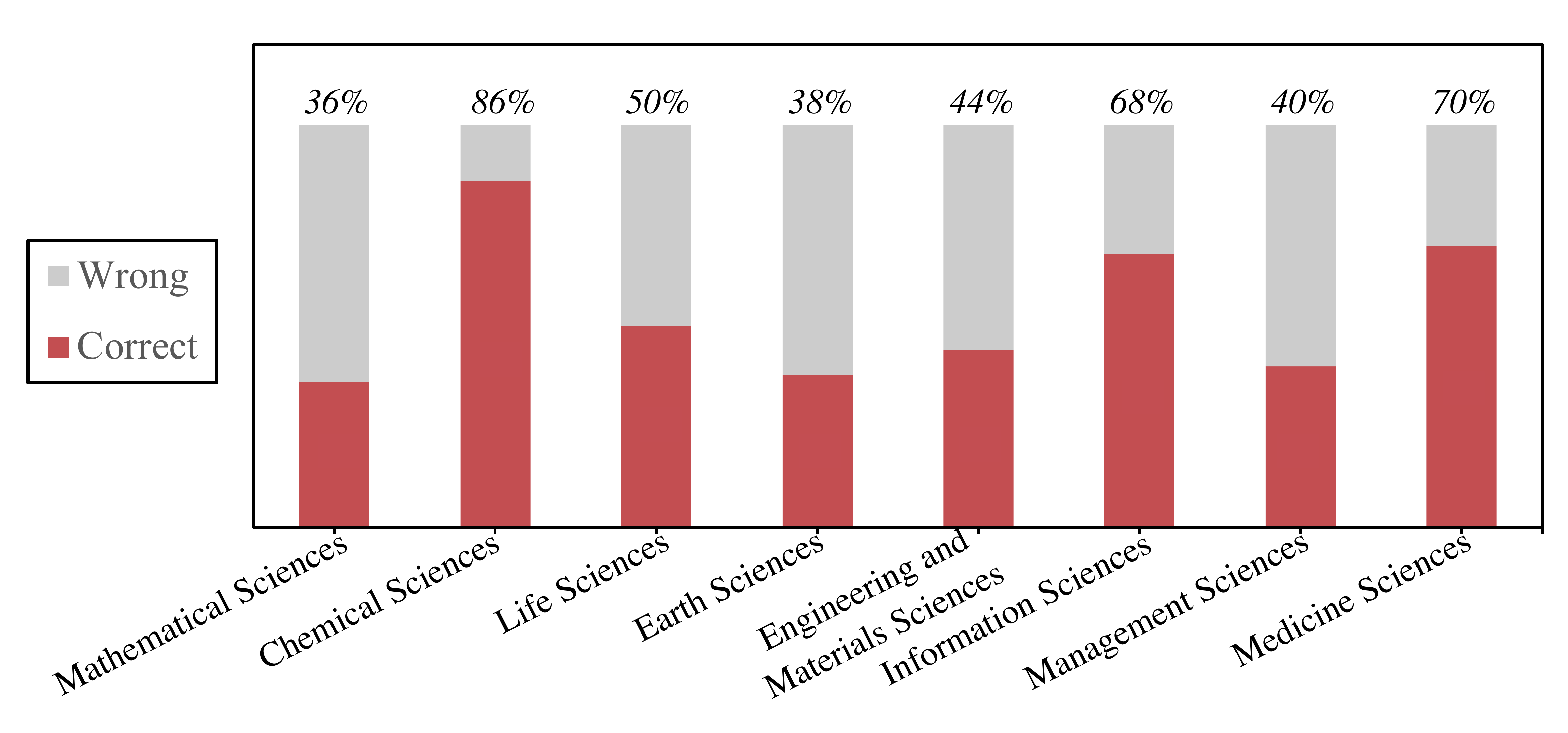} 
\vspace{-0.3cm}
\caption{The results of hidden interdiscipline find study.}
\label{Fig.evalresult}
\vspace{-0.5cm}
\end{figure}

Among all the wrong cases in RP-all, the HIRPCN classifies part of non-interdisciplinary research to interdisciplinary, which means except for their \textbf{labeled discipline}, our model predicts an \textbf{extra discipline}. 
% That may because the ground truth is wrong.
% 推测：在进行了错例观察后，我们发现这些额外预测的学科是与这些研究有一定的联系。
After observing those wrong cases, we noticed that these non-interdisciplinary research proposals were somewhat related to their extra predicted disciplines. 
To understand this phenomenon, We invited eight fund administrators from NSFC with an insightful understanding and management experience of the eight major disciplines to judge if those extra disciplines can provide helpful information for improving the reviewer assignment.
% 评价这些额外预测的学科信息能否用于更公平的审稿人分配
We first group those cases by the extra discipline predicted by HIRPCN. Then we sample 50 research proposals from each group. We send those research proposals to each expert according to the expert's familiar discipline. 
% 专家会标注这个额外的学科是否有用 yes no
If the expert thinks the reviewers from the extra discipline should be considered to supplement to the peer-review group, they will mark it as \textit{correct}, otherwise \textit{wrong}. We illustrated the percentage of \textit{correct} of each extra discipline in Figure~\ref{Fig.evalresult}. From the results, we can observe that:

(1) Overall, experts marked 57\% of the samples as \textit{correct}. Further, all experts comment that the HIRPCN can detect the hidden interdisciplinary research domain reasonably. 

% 有些专家评论，尽管有些case是错的，但他们在一定程度上展示了和predicted label的相关性，这些相关的内容不作为主要的研究内容，因此不是非常属于predicted label。这证明了我们的模型能够发现一些弱相关性。
(2) Some experts comment that although some cases are marked \textit{wrong}, most of them explicitly use the keywords from the predicted extra discipline.
Experts cannot categorize those cases into interdisciplinary research because their focal point is on the labeled discipline instead of the predicted extra one (e.g., a study on \textit{Artificial Intelligence} might introduce the idea from \textit{Neural Sciences}). 

% (2) Some experts comment that although some parts of the \textit{wrong} samples explicitly use the keywords from the predicted extra discipline. However, from their perspective, they cannot categorize them into interdisciplinary research due to their methodology or idea having little relation with the predicted discipline.  
% \note{revise the little}
% 

% 学科是进化的，有些学科是融合的。专家评论有些错例是因为，given的学科本身是交叉学科，且given的学科包含了predicted的学科，因此不需要给她指定这个predicted学科。例如，C1005 包含了A和B，而我们的模型预测出来B，
(3) Some experts comment that they marked some cases as \textit{wrong} because the discipline investigator has proposed the labeled discipline to represent an interdiscipline that covered the topic within the predicted discipline. Thus, the labeled discipline is enough to describe the research domain. For example, the discipline \textit{Bionics and Artificial Intelligent}, denoted by C1005 in \textit{Life Sciences}, represent the interdiscipline of \textit{Bionics} and \textit{Artificial Intelligent}. 

% (3) Some experts comment that they marked some cases as \textit{wrong} because the labeled discipline is more suitable to describe the research domain even if the given one and the predicted extra one were interdisciplinary in the past. The ApplyID is an evolving and dynamic discipline system. For example, there were no ApplyID codes represent the interdiscipline of \textit{Bionics} and \textit{Artificial Intelligent} in 2018. However, in 2020, there was a new discipline \textit{Bionics and Artificial Intelligent}, which is denoted by code \textit{C1005}, proposed to describe the related research field. 
% example -> 学科进化 讲清楚

We also post four research proposals from different disciplines, which our model predicts a domain from \textit{Information Sciences} as an extra result. The detail is discussed in Appendix~\ref{appendix:wrong}.

\vspace{-0.3cm}
\section{Related Works}
Our work is most related to deep-learning based hierarchical multi-label classification~\cite{cerri2015hierarchical} (HMC)  methods which leverage standard neural network approaches for multi-label classification problems~\cite{giunchiglia2021multi,wehrmann2018hierarchical,gargiulo2019deep} and then exploit the hierarchy constraint~\cite{huang2019hierarchical} in order to produce coherent predictions and improve performance. 
Zhang et al.~\cite{Zhang2021} propose a document categorization method with hierarchical structure under weak supervision. The work~\cite{tang2020multi} design a attention-based graph convolution network to category the patent to IPC codes. The work in \cite{LaGrassa2021} use the convolution neural network as an encoder and explore the HMC problem in image classification. The works in \cite{nakano2020active} propose a active learning approach for HMC problem. Aly et al.~\cite{Aly2019} propose a capsule network based method for HMC problem. Also there are some work focus on co-embedding the classes and entity into vector space for preserve the hierarchy structure. TAXOGAN~\cite{Yang2020} embedding the network nodes and hierarchical labels together, which focus on taxonomy modeling.
In this paper, we inherit the fundamental idea of HMC networks to construct our novel classification methods on the hierarchical discipline system.
\vspace{-0.3cm}\section{Conclusion}
This paper proposed HIRPCN, a novel HMC method for the IRPC task on the real-world research proposal dataset. This method extracts the semantic information from each document separately and combines them to achieve a high-level abstraction. Then, the model process each previous level's prediction result and neighborhood topology structure on the pre-defined interdisciplinary graph as interdisciplinary knowledge. The prediction step integrates each part of the research proposal by attention mechanism and generates the next level prediction. The experiments show that our model achieves the best performance on three real-world datasets and can provide the best granular on each level prediction. Other experiments explains the attention mechanism and points out that our model could fix the incomplete interdisciplinary labels under a domain-specific expert evaluation. With the ability to start prediction from any given label, HIRPCN could assist whether researcher or fund administrator to fill the discipline information, which is crucial to improve the reviewer assignation.
% We first propose a two-level hierarchical encoder to capture the representation of the proposal fully. Then, we use a Transformer-based decoder incorporating the previous label prediction and the proposal's representation to generate the label. Due to the characteristic of the decoder, our model can start from the root label or any given previous label, which corresponds to utilize the expert knowledge. We conduct extensive experiments to evaluate our model performance. The results show that our model achieves the best performance in the HMPC problem and is effective for the utilization of expert knowledge. Other experiments evaluate the ability of our model to give a proper granularity and to capture the label dependency.
% \vspace{-0.35cm}
\bibliographystyle{ACM-Reference-Format}
\bibliography{bib/reference}


\begin{thebibliography}{42}

%%% the Web address.

\ifx \showCODEN    \undefined \def \showCODEN     #1{\unskip}     \fi
\ifx \showDOI      \undefined \def \showDOI       #1{#1}\fi
\ifx \showISBNx    \undefined \def \showISBNx     #1{\unskip}     \fi
\ifx \showISBNxiii \undefined \def \showISBNxiii  #1{\unskip}     \fi
\ifx \showISSN     \undefined \def \showISSN      #1{\unskip}     \fi
\ifx \showLCCN     \undefined \def \showLCCN      #1{\unskip}     \fi
\ifx \shownote     \undefined \def \shownote      #1{#1}          \fi
\ifx \showarticletitle \undefined \def \showarticletitle #1{#1}   \fi
\ifx \showURL      \undefined \def \showURL       {\relax}        \fi
% invisible to TeX
\providecommand\bibfield[2]{#2}
\providecommand\bibinfo[2]{#2}
\providecommand\natexlab[1]{#1}
\providecommand\showeprint[2][]{arXiv:#2}

\bibitem[\protect\citeauthoryear{Aboelela, Larson, Bakken, Carrasquillo,
  Formicola, Glied, Haas, and Gebbie}{Aboelela et~al\mbox{.}}{2007}]%
        {aboelela2007defining}
\bibfield{author}{\bibinfo{person}{Sally~W Aboelela}, \bibinfo{person}{Elaine
  Larson}, \bibinfo{person}{Suzanne Bakken}, \bibinfo{person}{Olveen
  Carrasquillo}, \bibinfo{person}{Allan Formicola}, \bibinfo{person}{Sherry~A
  Glied}, \bibinfo{person}{Janet Haas}, {and} \bibinfo{person}{Kristine~M
  Gebbie}.} \bibinfo{year}{2007}\natexlab{}.
\newblock \showarticletitle{Defining interdisciplinary research: Conclusions
  from a critical review of the literature}.
\newblock \bibinfo{journal}{\emph{Health services research}}
  \bibinfo{volume}{42}, \bibinfo{number}{1p1} (\bibinfo{year}{2007}),
  \bibinfo{pages}{329--346}.
\newblock


\bibitem[\protect\citeauthoryear{Aly, Remus, and Biemann}{Aly
  et~al\mbox{.}}{2019}]%
        {Aly2019}
\bibfield{author}{\bibinfo{person}{Rami Aly}, \bibinfo{person}{Steffen Remus},
  {and} \bibinfo{person}{Chris Biemann}.} \bibinfo{year}{2019}\natexlab{}.
\newblock \showarticletitle{{Hierarchical multi-label classification of text
  with capsule networks}}.
\newblock \bibinfo{journal}{\emph{ACL 2019 - 57th Annual Meeting of the
  Association for Computational Linguistics, Proceedings of the Student
  Research Workshop}} (\bibinfo{year}{2019}), \bibinfo{pages}{323--330}.
\newblock
\showISBNx{9781950737475}
\urldef\tempurl%
\url{https://doi.org/10.18653/v1/p19-2045}
\showDOI{\tempurl}


\bibitem[\protect\citeauthoryear{Biggs, Schl{\"u}ter, Biggs, Bohensky,
  BurnSilver, Cundill, Dakos, Daw, Evans, Kotschy, et~al\mbox{.}}{Biggs
  et~al\mbox{.}}{2012}]%
        {biggs2012toward}
\bibfield{author}{\bibinfo{person}{Reinette Biggs}, \bibinfo{person}{Maja
  Schl{\"u}ter}, \bibinfo{person}{Duan Biggs}, \bibinfo{person}{Erin~L
  Bohensky}, \bibinfo{person}{Shauna BurnSilver}, \bibinfo{person}{Georgina
  Cundill}, \bibinfo{person}{Vasilis Dakos}, \bibinfo{person}{Tim~M Daw},
  \bibinfo{person}{Louisa~S Evans}, \bibinfo{person}{Karen Kotschy},
  {et~al\mbox{.}}} \bibinfo{year}{2012}\natexlab{}.
\newblock \showarticletitle{Toward principles for enhancing the resilience of
  ecosystem services}.
\newblock \bibinfo{journal}{\emph{Annual review of environment and resources}}
  \bibinfo{volume}{37} (\bibinfo{year}{2012}), \bibinfo{pages}{421--448}.
\newblock


\bibitem[\protect\citeauthoryear{Bojanowski, Grave, Joulin, and
  Mikolov}{Bojanowski et~al\mbox{.}}{2017a}]%
        {textcnn}
\bibfield{author}{\bibinfo{person}{Piotr Bojanowski}, \bibinfo{person}{Edouard
  Grave}, \bibinfo{person}{Armand Joulin}, {and} \bibinfo{person}{Tomas
  Mikolov}.} \bibinfo{year}{2017}\natexlab{a}.
\newblock \showarticletitle{{Enriching Word Vectors with Subword Information}}.
\newblock \bibinfo{journal}{\emph{Transactions of the Association for
  Computational Linguistics}}  \bibinfo{volume}{5} (\bibinfo{year}{2017}),
  \bibinfo{pages}{135--146}.
\newblock
\urldef\tempurl%
\url{https://doi.org/10.1162/tacl_a_00051}
\showDOI{\tempurl}
\showeprint[arxiv]{1607.04606}


\bibitem[\protect\citeauthoryear{Bojanowski, Grave, Joulin, and
  Mikolov}{Bojanowski et~al\mbox{.}}{2017b}]%
        {fasttext2}
\bibfield{author}{\bibinfo{person}{Piotr Bojanowski}, \bibinfo{person}{Edouard
  Grave}, \bibinfo{person}{Armand Joulin}, {and} \bibinfo{person}{Tomas
  Mikolov}.} \bibinfo{year}{2017}\natexlab{b}.
\newblock \showarticletitle{{Enriching Word Vectors with Subword Information}}.
\newblock \bibinfo{journal}{\emph{Transactions of the Association for
  Computational Linguistics}}  \bibinfo{volume}{5} (\bibinfo{year}{2017}),
  \bibinfo{pages}{135--146}.
\newblock
\showISSN{2307-387X}
\urldef\tempurl%
\url{https://doi.org/10.1162/tacl_a_00051}
\showDOI{\tempurl}
\showeprint[arxiv]{1607.04606}


\bibitem[\protect\citeauthoryear{Bojanowski, Grave, Joulin, and
  Mikolov}{Bojanowski et~al\mbox{.}}{2017c}]%
        {textrcnn}
\bibfield{author}{\bibinfo{person}{Piotr Bojanowski}, \bibinfo{person}{Edouard
  Grave}, \bibinfo{person}{Armand Joulin}, {and} \bibinfo{person}{Tomas
  Mikolov}.} \bibinfo{year}{2017}\natexlab{c}.
\newblock \showarticletitle{{Enriching Word Vectors with Subword Information}}.
\newblock \bibinfo{journal}{\emph{Transactions of the Association for
  Computational Linguistics}}  \bibinfo{volume}{5} (\bibinfo{year}{2017}),
  \bibinfo{pages}{135--146}.
\newblock
\showISSN{2307-387X}
\urldef\tempurl%
\url{https://doi.org/10.1162/tacl_a_00051}
\showDOI{\tempurl}
\showeprint[arxiv]{1607.04606}


\bibitem[\protect\citeauthoryear{Cerri, Barros, and de~Carvalho}{Cerri
  et~al\mbox{.}}{2015}]%
        {cerri2015hierarchical}
\bibfield{author}{\bibinfo{person}{Ricardo Cerri}, \bibinfo{person}{Rodrigo~C
  Barros}, {and} \bibinfo{person}{Andr{\'e}~CPLF de Carvalho}.}
  \bibinfo{year}{2015}\natexlab{}.
\newblock \showarticletitle{Hierarchical classification of gene ontology-based
  protein functions with neural networks}. In \bibinfo{booktitle}{\emph{2015
  international joint conference on neural networks (IJCNN)}}. IEEE,
  \bibinfo{pages}{1--8}.
\newblock


\bibitem[\protect\citeauthoryear{Dosovitskiy, Beyer, Kolesnikov, Weissenborn,
  Zhai, Unterthiner, Dehghani, Minderer, Heigold, Gelly,
  et~al\mbox{.}}{Dosovitskiy et~al\mbox{.}}{2020}]%
        {dosovitskiy2020image}
\bibfield{author}{\bibinfo{person}{Alexey Dosovitskiy}, \bibinfo{person}{Lucas
  Beyer}, \bibinfo{person}{Alexander Kolesnikov}, \bibinfo{person}{Dirk
  Weissenborn}, \bibinfo{person}{Xiaohua Zhai}, \bibinfo{person}{Thomas
  Unterthiner}, \bibinfo{person}{Mostafa Dehghani}, \bibinfo{person}{Matthias
  Minderer}, \bibinfo{person}{Georg Heigold}, \bibinfo{person}{Sylvain Gelly},
  {et~al\mbox{.}}} \bibinfo{year}{2020}\natexlab{}.
\newblock \showarticletitle{An image is worth 16x16 words: Transformers for
  image recognition at scale}.
\newblock \bibinfo{journal}{\emph{arXiv preprint arXiv:2010.11929}}
  (\bibinfo{year}{2020}).
\newblock


\bibitem[\protect\citeauthoryear{Fortunato, Bergstrom, B{\"o}rner, Evans,
  Helbing, Milojevi{\'c}, Petersen, Radicchi, Sinatra, Uzzi,
  et~al\mbox{.}}{Fortunato et~al\mbox{.}}{2018}]%
        {fortunato2018science}
\bibfield{author}{\bibinfo{person}{Santo Fortunato}, \bibinfo{person}{Carl~T
  Bergstrom}, \bibinfo{person}{Katy B{\"o}rner}, \bibinfo{person}{James~A
  Evans}, \bibinfo{person}{Dirk Helbing}, \bibinfo{person}{Sta{\v{s}}a
  Milojevi{\'c}}, \bibinfo{person}{Alexander~M Petersen},
  \bibinfo{person}{Filippo Radicchi}, \bibinfo{person}{Roberta Sinatra},
  \bibinfo{person}{Brian Uzzi}, {et~al\mbox{.}}}
  \bibinfo{year}{2018}\natexlab{}.
\newblock \showarticletitle{Science of science}.
\newblock \bibinfo{journal}{\emph{Science}} \bibinfo{volume}{359},
  \bibinfo{number}{6379} (\bibinfo{year}{2018}).
\newblock


\bibitem[\protect\citeauthoryear{Foster, Rzhetsky, and Evans}{Foster
  et~al\mbox{.}}{2015}]%
        {foster2015tradition}
\bibfield{author}{\bibinfo{person}{Jacob~G Foster}, \bibinfo{person}{Andrey
  Rzhetsky}, {and} \bibinfo{person}{James~A Evans}.}
  \bibinfo{year}{2015}\natexlab{}.
\newblock \showarticletitle{Tradition and innovation in scientists’ research
  strategies}.
\newblock \bibinfo{journal}{\emph{American Sociological Review}}
  \bibinfo{volume}{80}, \bibinfo{number}{5} (\bibinfo{year}{2015}),
  \bibinfo{pages}{875--908}.
\newblock


\bibitem[\protect\citeauthoryear{Gargiulo, Silvestri, Ciampi, and
  De~Pietro}{Gargiulo et~al\mbox{.}}{2019}]%
        {gargiulo2019deep}
\bibfield{author}{\bibinfo{person}{Francesco Gargiulo},
  \bibinfo{person}{Stefano Silvestri}, \bibinfo{person}{Mario Ciampi}, {and}
  \bibinfo{person}{Giuseppe De~Pietro}.} \bibinfo{year}{2019}\natexlab{}.
\newblock \showarticletitle{Deep neural network for hierarchical extreme
  multi-label text classification}.
\newblock \bibinfo{journal}{\emph{Applied Soft Computing}}
  \bibinfo{volume}{79} (\bibinfo{year}{2019}), \bibinfo{pages}{125--138}.
\newblock


\bibitem[\protect\citeauthoryear{Gibaja and Ventura}{Gibaja and
  Ventura}{2014}]%
        {metric1}
\bibfield{author}{\bibinfo{person}{Eva Gibaja} {and}
  \bibinfo{person}{Sebasti{\'a}n Ventura}.} \bibinfo{year}{2014}\natexlab{}.
\newblock \showarticletitle{Multi-label learning: a review of the state of the
  art and ongoing research}.
\newblock \bibinfo{journal}{\emph{Wiley Interdisciplinary Reviews: Data Mining
  and Knowledge Discovery}} \bibinfo{volume}{4}, \bibinfo{number}{6}
  (\bibinfo{year}{2014}), \bibinfo{pages}{411--444}.
\newblock


\bibitem[\protect\citeauthoryear{Giunchiglia and Lukasiewicz}{Giunchiglia and
  Lukasiewicz}{2021}]%
        {giunchiglia2021multi}
\bibfield{author}{\bibinfo{person}{Eleonora Giunchiglia} {and}
  \bibinfo{person}{Thomas Lukasiewicz}.} \bibinfo{year}{2021}\natexlab{}.
\newblock \showarticletitle{Multi-Label Classification Neural Networks with
  Hard Logical Constraints}.
\newblock \bibinfo{journal}{\emph{arXiv preprint arXiv:2103.13427}}
  (\bibinfo{year}{2021}).
\newblock


\bibitem[\protect\citeauthoryear{Han, Xiao, Wu, Guo, Xu, and Wang}{Han
  et~al\mbox{.}}{2021}]%
        {han2021transformer}
\bibfield{author}{\bibinfo{person}{Kai Han}, \bibinfo{person}{An Xiao},
  \bibinfo{person}{Enhua Wu}, \bibinfo{person}{Jianyuan Guo},
  \bibinfo{person}{Chunjing Xu}, {and} \bibinfo{person}{Yunhe Wang}.}
  \bibinfo{year}{2021}\natexlab{}.
\newblock \showarticletitle{Transformer in transformer}.
\newblock \bibinfo{journal}{\emph{arXiv preprint arXiv:2103.00112}}
  (\bibinfo{year}{2021}).
\newblock


\bibitem[\protect\citeauthoryear{He}{He}{1999}]%
        {he1999knowledge}
\bibfield{author}{\bibinfo{person}{Qin He}.} \bibinfo{year}{1999}\natexlab{}.
\newblock \showarticletitle{Knowledge discovery through co-word analysis}.
\newblock \bibinfo{journal}{\emph{Library Trends}} \bibinfo{volume}{48},
  \bibinfo{number}{1} (\bibinfo{year}{1999}), \bibinfo{pages}{133--159}.
\newblock
\newblock
\shownote{Retrieved from http://eric.ed.gov/?id=EJ595487.}


\bibitem[\protect\citeauthoryear{Huang, Chen, Liu, Chen, Huang, Liu, Zhao,
  Zhang, and Wang}{Huang et~al\mbox{.}}{2019}]%
        {huang2019hierarchical}
\bibfield{author}{\bibinfo{person}{Wei Huang}, \bibinfo{person}{Enhong Chen},
  \bibinfo{person}{Qi Liu}, \bibinfo{person}{Yuying Chen}, \bibinfo{person}{Zai
  Huang}, \bibinfo{person}{Yang Liu}, \bibinfo{person}{Zhou Zhao},
  \bibinfo{person}{Dan Zhang}, {and} \bibinfo{person}{Shijin Wang}.}
  \bibinfo{year}{2019}\natexlab{}.
\newblock \showarticletitle{Hierarchical multi-label text classification: An
  attention-based recurrent network approach}. In
  \bibinfo{booktitle}{\emph{Proceedings of the 28th ACM International
  Conference on Information and Knowledge Management}}.
  \bibinfo{pages}{1051--1060}.
\newblock


\bibitem[\protect\citeauthoryear{Johnson and Zhang}{Johnson and Zhang}{2017}]%
        {dpcnn}
\bibfield{author}{\bibinfo{person}{Rie Johnson} {and} \bibinfo{person}{Tong
  Zhang}.} \bibinfo{year}{2017}\natexlab{}.
\newblock \showarticletitle{{Deep pyramid convolutional neural networks for
  text categorization}}.
\newblock \bibinfo{journal}{\emph{ACL 2017 - 55th Annual Meeting of the
  Association for Computational Linguistics, Proceedings of the Conference
  (Long Papers)}}  \bibinfo{volume}{1} (\bibinfo{year}{2017}),
  \bibinfo{pages}{562--570}.
\newblock
\showISBNx{9781945626753}
\urldef\tempurl%
\url{https://doi.org/10.18653/v1/P17-1052}
\showDOI{\tempurl}


\bibitem[\protect\citeauthoryear{Joulin, Grave, Bojanowski, and Mikolov}{Joulin
  et~al\mbox{.}}{2017}]%
        {fasttext1}
\bibfield{author}{\bibinfo{person}{Armand Joulin}, \bibinfo{person}{Edouard
  Grave}, \bibinfo{person}{Piotr Bojanowski}, {and} \bibinfo{person}{Tomas
  Mikolov}.} \bibinfo{year}{2017}\natexlab{}.
\newblock \showarticletitle{{Bag of tricks for efficient text classification}}.
\newblock \bibinfo{journal}{\emph{15th Conference of the European Chapter of
  the Association for Computational Linguistics, EACL 2017 - Proceedings of
  Conference}}  \bibinfo{volume}{2} (\bibinfo{year}{2017}),
  \bibinfo{pages}{427--431}.
\newblock
\showISBNx{9781510838604}
\urldef\tempurl%
\url{https://doi.org/10.18653/v1/e17-2068}
\showDOI{\tempurl}
\showeprint[arxiv]{1607.01759}


\bibitem[\protect\citeauthoryear{Kingma and Ba}{Kingma and Ba}{2014}]%
        {kingma2014adam}
\bibfield{author}{\bibinfo{person}{Diederik~P Kingma} {and}
  \bibinfo{person}{Jimmy Ba}.} \bibinfo{year}{2014}\natexlab{}.
\newblock \showarticletitle{Adam: A method for stochastic optimization}.
\newblock \bibinfo{journal}{\emph{arXiv preprint arXiv:1412.6980}}
  (\bibinfo{year}{2014}).
\newblock


\bibitem[\protect\citeauthoryear{Kipf and Welling}{Kipf and Welling}{2016}]%
        {kipf2016semi}
\bibfield{author}{\bibinfo{person}{Thomas~N Kipf} {and} \bibinfo{person}{Max
  Welling}.} \bibinfo{year}{2016}\natexlab{}.
\newblock \showarticletitle{Semi-supervised classification with graph
  convolutional networks}.
\newblock \bibinfo{journal}{\emph{arXiv preprint arXiv:1609.02907}}
  (\bibinfo{year}{2016}).
\newblock


\bibitem[\protect\citeauthoryear{{La Grassa}, Gallo, and Landro}{{La Grassa}
  et~al\mbox{.}}{2021}]%
        {LaGrassa2021}
\bibfield{author}{\bibinfo{person}{Riccardo {La Grassa}},
  \bibinfo{person}{Ignazio Gallo}, {and} \bibinfo{person}{Nicola Landro}.}
  \bibinfo{year}{2021}\natexlab{}.
\newblock \showarticletitle{{Learn class hierarchy using convolutional neural
  networks}}.
\newblock \bibinfo{journal}{\emph{Applied Intelligence}}
  (\bibinfo{year}{2021}), \bibinfo{pages}{1--7}.
\newblock
\showISSN{15737497}
\urldef\tempurl%
\url{https://doi.org/10.1007/s10489-020-02103-6}
\showDOI{\tempurl}
\showeprint[arxiv]{2005.08622}


\bibitem[\protect\citeauthoryear{Liu, Qiu, and Xuanjing}{Liu
  et~al\mbox{.}}{2016}]%
        {textrnn}
\bibfield{author}{\bibinfo{person}{Pengfei Liu}, \bibinfo{person}{Xipeng Qiu},
  {and} \bibinfo{person}{Huang Xuanjing}.} \bibinfo{year}{2016}\natexlab{}.
\newblock \showarticletitle{{Recurrent neural network for text classification
  with multi-task learning}}.
\newblock \bibinfo{journal}{\emph{IJCAI International Joint Conference on
  Artificial Intelligence}}  \bibinfo{volume}{2016-Janua}
  (\bibinfo{year}{2016}), \bibinfo{pages}{2873--2879}.
\newblock
\showISSN{10450823}
\showeprint[arxiv]{1605.05101}


\bibitem[\protect\citeauthoryear{Liu and Lapata}{Liu and Lapata}{2019}]%
        {liu2019hierarchical}
\bibfield{author}{\bibinfo{person}{Yang Liu} {and} \bibinfo{person}{Mirella
  Lapata}.} \bibinfo{year}{2019}\natexlab{}.
\newblock \showarticletitle{Hierarchical transformers for multi-document
  summarization}.
\newblock \bibinfo{journal}{\emph{arXiv preprint arXiv:1905.13164}}
  (\bibinfo{year}{2019}).
\newblock


\bibitem[\protect\citeauthoryear{{Mikolov}, {Sutskever}, {Chen}, {Corrado}, and
  {Dean}}{{Mikolov} et~al\mbox{.}}{2013}]%
        {mikolov2013distributed}
\bibfield{author}{\bibinfo{person}{Tomas {Mikolov}}, \bibinfo{person}{Ilya
  {Sutskever}}, \bibinfo{person}{Kai {Chen}}, \bibinfo{person}{Greg~S
  {Corrado}}, {and} \bibinfo{person}{Jeff {Dean}}.}
  \bibinfo{year}{2013}\natexlab{}.
\newblock \showarticletitle{Distributed Representations of Words and Phrases
  and their Compositionality}. In \bibinfo{booktitle}{\emph{Advances in Neural
  Information Processing Systems 26}}, Vol.~\bibinfo{volume}{26}.
  \bibinfo{pages}{3111--3119}.
\newblock


\bibitem[\protect\citeauthoryear{Milojevi{\'c}}{Milojevi{\'c}}{2015}]%
        {milojevic2015}
\bibfield{author}{\bibinfo{person}{Sta{\v{s}}a Milojevi{\'c}}.}
  \bibinfo{year}{2015}\natexlab{}.
\newblock \showarticletitle{Quantifying the cognitive extent of science}.
\newblock \bibinfo{journal}{\emph{Journal of Informetrics}}
  \bibinfo{volume}{9}, \bibinfo{number}{4} (\bibinfo{year}{2015}),
  \bibinfo{pages}{962--973}.
\newblock


\bibitem[\protect\citeauthoryear{Nakano, Cerri, and Vens}{Nakano
  et~al\mbox{.}}{2020}]%
        {nakano2020active}
\bibfield{author}{\bibinfo{person}{Felipe~Kenji Nakano},
  \bibinfo{person}{Ricardo Cerri}, {and} \bibinfo{person}{Celine Vens}.}
  \bibinfo{year}{2020}\natexlab{}.
\newblock \showarticletitle{Active learning for hierarchical multi-label
  classification}.
\newblock \bibinfo{journal}{\emph{Data Mining and Knowledge Discovery}}
  \bibinfo{volume}{34}, \bibinfo{number}{5} (\bibinfo{year}{2020}),
  \bibinfo{pages}{1496--1530}.
\newblock


\bibitem[\protect\citeauthoryear{Pappagari, Zelasko, Villalba, Carmiel, and
  Dehak}{Pappagari et~al\mbox{.}}{2019}]%
        {pappagari2019hierarchical}
\bibfield{author}{\bibinfo{person}{Raghavendra Pappagari},
  \bibinfo{person}{Piotr Zelasko}, \bibinfo{person}{Jes{\'u}s Villalba},
  \bibinfo{person}{Yishay Carmiel}, {and} \bibinfo{person}{Najim Dehak}.}
  \bibinfo{year}{2019}\natexlab{}.
\newblock \showarticletitle{Hierarchical transformers for long document
  classification}. In \bibinfo{booktitle}{\emph{2019 IEEE Automatic Speech
  Recognition and Understanding Workshop (ASRU)}}. IEEE,
  \bibinfo{pages}{838--844}.
\newblock


\bibitem[\protect\citeauthoryear{Paszke, Gross, Massa, Lerer, Bradbury, Chanan,
  Killeen, Lin, Gimelshein, Antiga, et~al\mbox{.}}{Paszke
  et~al\mbox{.}}{2019}]%
        {paszke2019pytorch}
\bibfield{author}{\bibinfo{person}{Adam Paszke}, \bibinfo{person}{Sam Gross},
  \bibinfo{person}{Francisco Massa}, \bibinfo{person}{Adam Lerer},
  \bibinfo{person}{James Bradbury}, \bibinfo{person}{Gregory Chanan},
  \bibinfo{person}{Trevor Killeen}, \bibinfo{person}{Zeming Lin},
  \bibinfo{person}{Natalia Gimelshein}, \bibinfo{person}{Luca Antiga},
  {et~al\mbox{.}}} \bibinfo{year}{2019}\natexlab{}.
\newblock \showarticletitle{Pytorch: An imperative style, high-performance deep
  learning library}.
\newblock \bibinfo{journal}{\emph{arXiv preprint arXiv:1912.01703}}
  (\bibinfo{year}{2019}).
\newblock


\bibitem[\protect\citeauthoryear{Porter and Rafols}{Porter and Rafols}{2009}]%
        {porter2009science}
\bibfield{author}{\bibinfo{person}{Alan Porter} {and} \bibinfo{person}{Ismael
  Rafols}.} \bibinfo{year}{2009}\natexlab{}.
\newblock \showarticletitle{Is science becoming more interdisciplinary?
  Measuring and mapping six research fields over time}.
\newblock \bibinfo{journal}{\emph{Scientometrics}} \bibinfo{volume}{81},
  \bibinfo{number}{3} (\bibinfo{year}{2009}), \bibinfo{pages}{719--745}.
\newblock


\bibitem[\protect\citeauthoryear{Rafols and Meyer}{Rafols and Meyer}{2010}]%
        {rafols2010diversity}
\bibfield{author}{\bibinfo{person}{Ismael Rafols} {and} \bibinfo{person}{Martin
  Meyer}.} \bibinfo{year}{2010}\natexlab{}.
\newblock \showarticletitle{Diversity and network coherence as indicators of
  interdisciplinarity: case studies in bionanoscience}.
\newblock \bibinfo{journal}{\emph{Scientometrics}} \bibinfo{volume}{82},
  \bibinfo{number}{2} (\bibinfo{year}{2010}), \bibinfo{pages}{263--287}.
\newblock


\bibitem[\protect\citeauthoryear{Stirling}{Stirling}{2007}]%
        {stirling2007general}
\bibfield{author}{\bibinfo{person}{Andy Stirling}.}
  \bibinfo{year}{2007}\natexlab{}.
\newblock \showarticletitle{A general framework for analysing diversity in
  science, technology and society}.
\newblock \bibinfo{journal}{\emph{Journal of the Royal Society Interface}}
  \bibinfo{volume}{4}, \bibinfo{number}{15} (\bibinfo{year}{2007}),
  \bibinfo{pages}{707--719}.
\newblock


\bibitem[\protect\citeauthoryear{Tang, Jiang, Xia, Pitera, Welser, and
  Chawla}{Tang et~al\mbox{.}}{2020}]%
        {tang2020multi}
\bibfield{author}{\bibinfo{person}{Pingjie Tang}, \bibinfo{person}{Meng Jiang},
  \bibinfo{person}{Bryan~Ning Xia}, \bibinfo{person}{Jed~W Pitera},
  \bibinfo{person}{Jeffrey Welser}, {and} \bibinfo{person}{Nitesh~V Chawla}.}
  \bibinfo{year}{2020}\natexlab{}.
\newblock \showarticletitle{Multi-label patent categorization with non-local
  attention-based graph convolutional network}. In
  \bibinfo{booktitle}{\emph{Proceedings of the AAAI Conference on Artificial
  Intelligence}}, Vol.~\bibinfo{volume}{34}. \bibinfo{pages}{9024--9031}.
\newblock


\bibitem[\protect\citeauthoryear{{Vaswani}, {Shazeer}, {Parmar}, {Uszkoreit},
  {Jones}, {Gomez}, {Kaiser}, and {Polosukhin}}{{Vaswani}
  et~al\mbox{.}}{2017}]%
        {vaswani2017attention}
\bibfield{author}{\bibinfo{person}{Ashish {Vaswani}}, \bibinfo{person}{Noam
  {Shazeer}}, \bibinfo{person}{Niki {Parmar}}, \bibinfo{person}{Jakob
  {Uszkoreit}}, \bibinfo{person}{Llion {Jones}}, \bibinfo{person}{Aidan~N.
  {Gomez}}, \bibinfo{person}{Lukasz {Kaiser}}, {and} \bibinfo{person}{Illia
  {Polosukhin}}.} \bibinfo{year}{2017}\natexlab{}.
\newblock \showarticletitle{Attention is All You Need}. In
  \bibinfo{booktitle}{\emph{Proceedings of the 31st International Conference on
  Neural Information Processing Systems}}, Vol.~\bibinfo{volume}{30}.
  \bibinfo{pages}{5998--6008}.
\newblock


\bibitem[\protect\citeauthoryear{Vens, Struyf, Schietgat, D{\v{z}}eroski, and
  Blockeel}{Vens et~al\mbox{.}}{2008}]%
        {metric2}
\bibfield{author}{\bibinfo{person}{Celine Vens}, \bibinfo{person}{Jan Struyf},
  \bibinfo{person}{Leander Schietgat}, \bibinfo{person}{Sa{\v{s}}o
  D{\v{z}}eroski}, {and} \bibinfo{person}{Hendrik Blockeel}.}
  \bibinfo{year}{2008}\natexlab{}.
\newblock \showarticletitle{Decision trees for hierarchical multi-label
  classification}.
\newblock \bibinfo{journal}{\emph{Machine learning}} \bibinfo{volume}{73},
  \bibinfo{number}{2} (\bibinfo{year}{2008}), \bibinfo{pages}{185}.
\newblock


\bibitem[\protect\citeauthoryear{Wehrmann, Cerri, and Barros}{Wehrmann
  et~al\mbox{.}}{2018}]%
        {wehrmann2018hierarchical}
\bibfield{author}{\bibinfo{person}{Jonatas Wehrmann}, \bibinfo{person}{Ricardo
  Cerri}, {and} \bibinfo{person}{Rodrigo Barros}.}
  \bibinfo{year}{2018}\natexlab{}.
\newblock \showarticletitle{Hierarchical multi-label classification networks}.
  In \bibinfo{booktitle}{\emph{International Conference on Machine Learning}}.
  PMLR, \bibinfo{pages}{5075--5084}.
\newblock


\bibitem[\protect\citeauthoryear{Winzer}{Winzer}{2012}]%
        {winzer2012conceptualizing}
\bibfield{author}{\bibinfo{person}{Christian Winzer}.}
  \bibinfo{year}{2012}\natexlab{}.
\newblock \showarticletitle{Conceptualizing energy security}.
\newblock \bibinfo{journal}{\emph{Energy policy}}  \bibinfo{volume}{46}
  (\bibinfo{year}{2012}), \bibinfo{pages}{36--48}.
\newblock


\bibitem[\protect\citeauthoryear{Wu, Zhang, and Honavar}{Wu
  et~al\mbox{.}}{2005}]%
        {wu2005learning}
\bibfield{author}{\bibinfo{person}{Feihong Wu}, \bibinfo{person}{Jun Zhang},
  {and} \bibinfo{person}{Vasant Honavar}.} \bibinfo{year}{2005}\natexlab{}.
\newblock \showarticletitle{Learning classifiers using hierarchically
  structured class taxonomies}. In \bibinfo{booktitle}{\emph{International
  symposium on abstraction, reformulation, and approximation}}. Springer,
  \bibinfo{pages}{313--320}.
\newblock


\bibitem[\protect\citeauthoryear{Xiao, Qiao, Fu, Du, and Wang}{Xiao
  et~al\mbox{.}}{2021}]%
        {xiao2021expert}
\bibfield{author}{\bibinfo{person}{Meng Xiao}, \bibinfo{person}{Ziyue Qiao},
  \bibinfo{person}{Yanjie Fu}, \bibinfo{person}{Yi Du}, {and}
  \bibinfo{person}{Pengyang Wang}.} \bibinfo{year}{2021}\natexlab{}.
\newblock \showarticletitle{Expert Knowledge-Guided Length-Variant Hierarchical
  Label Generation for Proposal Classification}.
\newblock \bibinfo{journal}{\emph{2021 IEEE International Conference on Data
  Mining}} (\bibinfo{year}{2021}), \bibinfo{pages}{757--766}.
\newblock


\bibitem[\protect\citeauthoryear{Yang, Zhang, and Han}{Yang
  et~al\mbox{.}}{2020}]%
        {Yang2020}
\bibfield{author}{\bibinfo{person}{Carl Yang}, \bibinfo{person}{Jieyu Zhang},
  {and} \bibinfo{person}{Jiawei Han}.} \bibinfo{year}{2020}\natexlab{}.
\newblock \showarticletitle{{Co-embedding network nodes and hierarchical labels
  with taxonomy based generative adversarial networks}}.
\newblock \bibinfo{journal}{\emph{Proceedings - IEEE International Conference
  on Data Mining, ICDM}}  \bibinfo{volume}{2020-Novem} (\bibinfo{year}{2020}),
  \bibinfo{pages}{721--730}.
\newblock
\showISBNx{9781728183169}
\showISSN{15504786}
\urldef\tempurl%
\url{https://doi.org/10.1109/ICDM50108.2020.00081}
\showDOI{\tempurl}


\bibitem[\protect\citeauthoryear{Zhang, Wei, and Zhou}{Zhang
  et~al\mbox{.}}{2019}]%
        {zhang2019hibert}
\bibfield{author}{\bibinfo{person}{Xingxing Zhang}, \bibinfo{person}{Furu Wei},
  {and} \bibinfo{person}{Ming Zhou}.} \bibinfo{year}{2019}\natexlab{}.
\newblock \showarticletitle{HIBERT: Document level pre-training of hierarchical
  bidirectional transformers for document summarization}.
\newblock \bibinfo{journal}{\emph{arXiv preprint arXiv:1905.06566}}
  (\bibinfo{year}{2019}).
\newblock


\bibitem[\protect\citeauthoryear{Zhang, Chen, Meng, and Han}{Zhang
  et~al\mbox{.}}{2021}]%
        {Zhang2021}
\bibfield{author}{\bibinfo{person}{Yu Zhang}, \bibinfo{person}{Xiusi Chen},
  \bibinfo{person}{Yu Meng}, {and} \bibinfo{person}{Jiawei Han}.}
  \bibinfo{year}{2021}\natexlab{}.
\newblock \showarticletitle{Hierarchical Metadata-Aware Document Categorization
  under Weak Supervision}. In \bibinfo{booktitle}{\emph{Proceedings of the 14th
  ACM International Conference on Web Search and Data Mining}}.
  \bibinfo{pages}{770--778}.
\newblock


\bibitem[\protect\citeauthoryear{Zhou, Shi, Tian, Qi, Li, Hao, and Xu}{Zhou
  et~al\mbox{.}}{2016}]%
        {textrnnattn}
\bibfield{author}{\bibinfo{person}{Peng Zhou}, \bibinfo{person}{Wei Shi},
  \bibinfo{person}{Jun Tian}, \bibinfo{person}{Zhenyu Qi},
  \bibinfo{person}{Bingchen Li}, \bibinfo{person}{Hongwei Hao}, {and}
  \bibinfo{person}{Bo Xu}.} \bibinfo{year}{2016}\natexlab{}.
\newblock \showarticletitle{{Attention-based bidirectional long short-term
  memory networks for relation classification}}.
\newblock \bibinfo{journal}{\emph{54th Annual Meeting of the Association for
  Computational Linguistics, ACL 2016 - Short Papers}} (\bibinfo{year}{2016}),
  \bibinfo{pages}{207--212}.
\newblock
\showISBNx{9781510827592}
\urldef\tempurl%
\url{https://doi.org/10.18653/v1/p16-2034}
\showDOI{\tempurl}


\end{thebibliography}
\clearpage
\appendix
\section{Appendix}
% For model implementation, we introduce the Transformer based model proposed in \cite{vaswani2017attention} and the Graph Convolution Network proposed in \cite{kipf2016semi}, which are both two state-of-the-art approaches and have demonstrated very competitive ability in combination with encoder-decoder based textual related models~\cite{devlin2018bert, vaswani2017attention} and structural related classification models~\cite{Chen2017, Anderson2018, peng2019hierarchical, zhou2020hierarchy}. 

\subsection{Graph Convolutional Networks} \label{appendix.gcn}
We adopt Graph Convolutional Networks (GCNs)~\cite{kipf2016semi} to extract the network-structured interdisciplinary knowledge from the Co-topic graph. 
GCN is the most popular and widely-used graph neural networks (GNNs), which is a deep graph information extractor that integrates node features and its local neighbors' information into low-dimensional representation vectors.

Formally, given a graph $G$, suppose $\mathcal{W}\in \mathbb{R}^{N \times N}$ is the adjacency matrix of graph topology, $\mathcal{H}^{0}\in \mathbb{R}^{N \times h}$ is the node feature matrix, $N$ is the total number of nodes, and $h$ is the dimension of node feature.
GCN uses an efficient layer-wise propagation rule based on a first-order approximation of spectral convolutions on graphs. 
The $l$-th GCN layer can be formulated as:

\begin{equation}
\begin{aligned}
    \mathcal{H}^{(l+1)} &= GCNlayer(\mathcal{W},\mathcal{H}^{(l)})\\
    &= \sigma(\tilde{\mathcal{W}}\mathcal{H}^{(l)}W^{(l)})
    %&X =H^{(0)}\\
    %&H^{(l+1)} = \sigma(\tilde{D}^{-\frac{1}{2}}\tilde{A}\tilde{D}^{-\frac{1}{2}}H^{(l)}W^{(l)})\\
    %&Z =H^{(L)}
\end{aligned}
\end{equation}
where $\mathcal{H}^{(l+1)}\in \mathbb{R}^{N \times h}$ is the output of the GCN layer, $W^{(l)}\in \mathbb{R}^{h \times h}$ is the layer-specific parameter matrix, $h$ is the hidden dimension, $\sigma$ denotes an activation function. $\tilde{\mathcal{W}}= \tilde{D}^{-\frac{1}{2}}\bar{\mathcal{W}}\tilde{D}^{-\frac{1}{2}}$, $\bar{\mathcal{W}} = A+I_N$, $I_N$ is the identity matrix and $\tilde{D}$ is the diagonal degree matrix of $\bar{\mathcal{W}}$.

In summary, after a $N$-layer GCN's propagation, formulated as $\mathcal{H}^{(N)} = N \times  GCNlayer(\mathcal{W},\mathcal{H}^{(0)})$, we can obtain the final node embedding matrix $\mathcal{H}^{(N)}\in \mathbb{R}^{N \times h}$, where each node embedding is integrated with the its $N$-hops neighborhood information.

\subsection{Transformer Details} \label{appendix.transformer}
A Transformer~\cite{vaswani2017attention} model usually has multiple layers. A layer of transformer model (i.e, a Transformer block) consists of a \textit{Multi-Head Self-Attention Layer}, a \textit{Residual Connections and Layer Normalization Layer}, a \textit{Feed Forward Layer}, and a \textit{Residual Connections and  Normalization Layer}, which can be written as:

\begin{equation}
\label{eq:3}
    Z^{(l)} = LN(X^{(l)}+MultiHead(X^{(l)},X^{(l)},X^{(l)})),
\end{equation}

\begin{equation}
\label{eq:4}
    X^{(l+1)} =  LN(Z^{(l)}+FC((Z^{(l)})),
\end{equation}

where $X^{(l)} = [x_1^{(l)}, x_2^{(l)}, ..., x_s^{(l)}]$ is the input sequence of $l$-th layer of Transformer, $s$ is the length of input sequence, $ x_i^{(l)}\in \mathbb{R}^h$ and $h$ in the dimension. $LN(\cdot)$ is layer normalization, $FC(\cdot)$ denotes a two-layer feed-forward network with $ReLU$ activation function, and $MultiHead(\cdot)$ denotes the multi-head attention layer, which is
calculated as follows:

\begin{equation}
\label{eq5}
MultiHead(Q, K, V) = Concat (head_1, . . . , head_h)W^O,
\end{equation}

\begin{equation}
head_i = Attention(Q W_i^Q,K W_i^K,V W_i^V),
\end{equation}

\begin{equation}
\label{eq7}
Attention (Q,K,V) = softmax(\frac{QK^T}{\sqrt{d}})V,
\end{equation}

where $W_i^Q, W_i^K, W_i^V\in \mathbb{R}^{h\times h}$ are weight matrices and $d$ is the number of attention heads. After the attention calculation, the $h$ outputs are concatenated and
transformed using a output weight matrix $W^O\in \mathbb{R}^{dh\times h}$.

In summary, given the input token sequence $X = [x_1, x_2, ..., x_s]$, we first initilize the input matrix  $X^{(0)} = [x_1^{(0)}, x_2^{(0)}, ..., x_s^{(0)}]$ before the first layer by a look-up table. Then, after the propagation in Equation \ref{eq:3} and \ref{eq:4} on a $N$-layers Transformer, formulized as $X^{(N)} = N \times Transformer(X^{(0)})$, we can obtain the final output embeddings of this sequence $X^{(N)} = [x_1^{(N)}, x_2^{(N)}, ..., x_s^{(N)}]$, where each embedding has contained the context information in this sequence.  The main hyper-parameters of a Transformer are the number of layers (i.e., Transformer blocks), the number of self-attention heads, and the maximum length of inputs.

\subsection{Hyperparameter Selection}
\begin{figure}[t!h]
\centering
\captionsetup{justification=centering}
\includegraphics[width=0.40\textwidth]{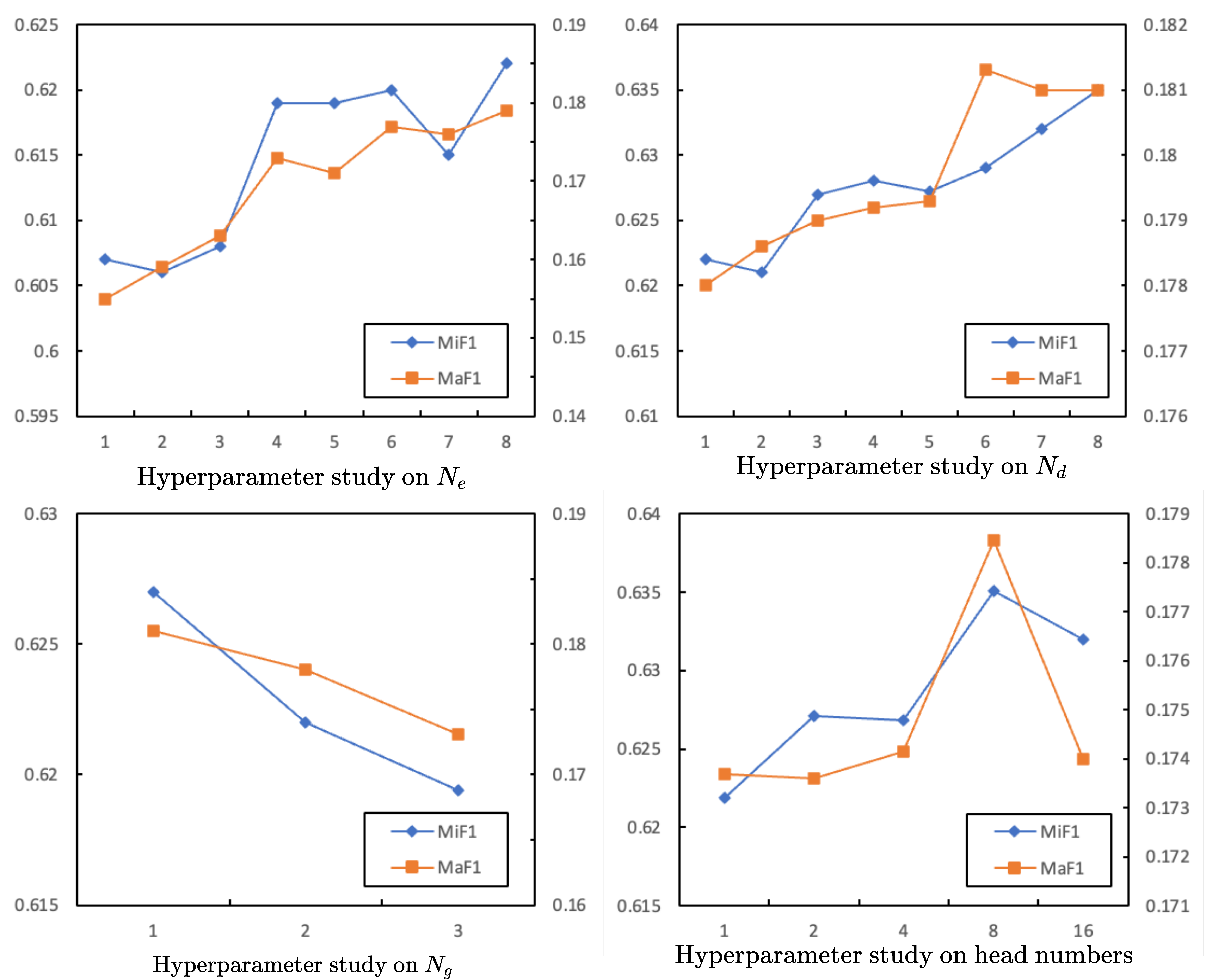} 
\caption{Hyper Parameter Selection Study.}
\vspace{-0.3cm}
\label{Fig.hyper}
\end{figure}

We conduct experiments to evaluate the effect of four key hyperparameters, i.e., $N_e$, $N_d$, $N_g$, and the number of self-attention heads in the SIE and IF. We investigate the sensitivity of those parameters in each part of HIRPCN and report the results in Figure \ref{Fig.hyper}. From the figure, we can observe that: 
(1) When the $N_e$ and $N_d$ varies from 1 to 8, the performance of  HIRPCN increases at first then decreases slowly. With more parameters introduced, the model will be likely over-fitting to impact the classification results.
(2) We found that the model would have the best performance when IKE sampled a one-hop neighborhood to extract interdisciplinary knowledge. The reason is that the interdisciplinary graph has many edges, and two or more hop sampling strategies will make the sampled sub-graph denser, which will provide unrelated knowledge for IKE.
(3) More attention heads will generally improve the performance of  HIRPCN. Meanwhile, we find that too many attention heads will make the model perform worse.
From the experimental result, we set $N_e$ to 8, $N_d$ to 8, $N_g$ to 1, and the number of self-attention heads to 8 to achieve the best performance. 

\subsection{Other Attention Study} \label{appendix:attention}
\begin{figure}[t!h]
\centering
\captionsetup{justification=centering}
\includegraphics[width=0.45\textwidth]{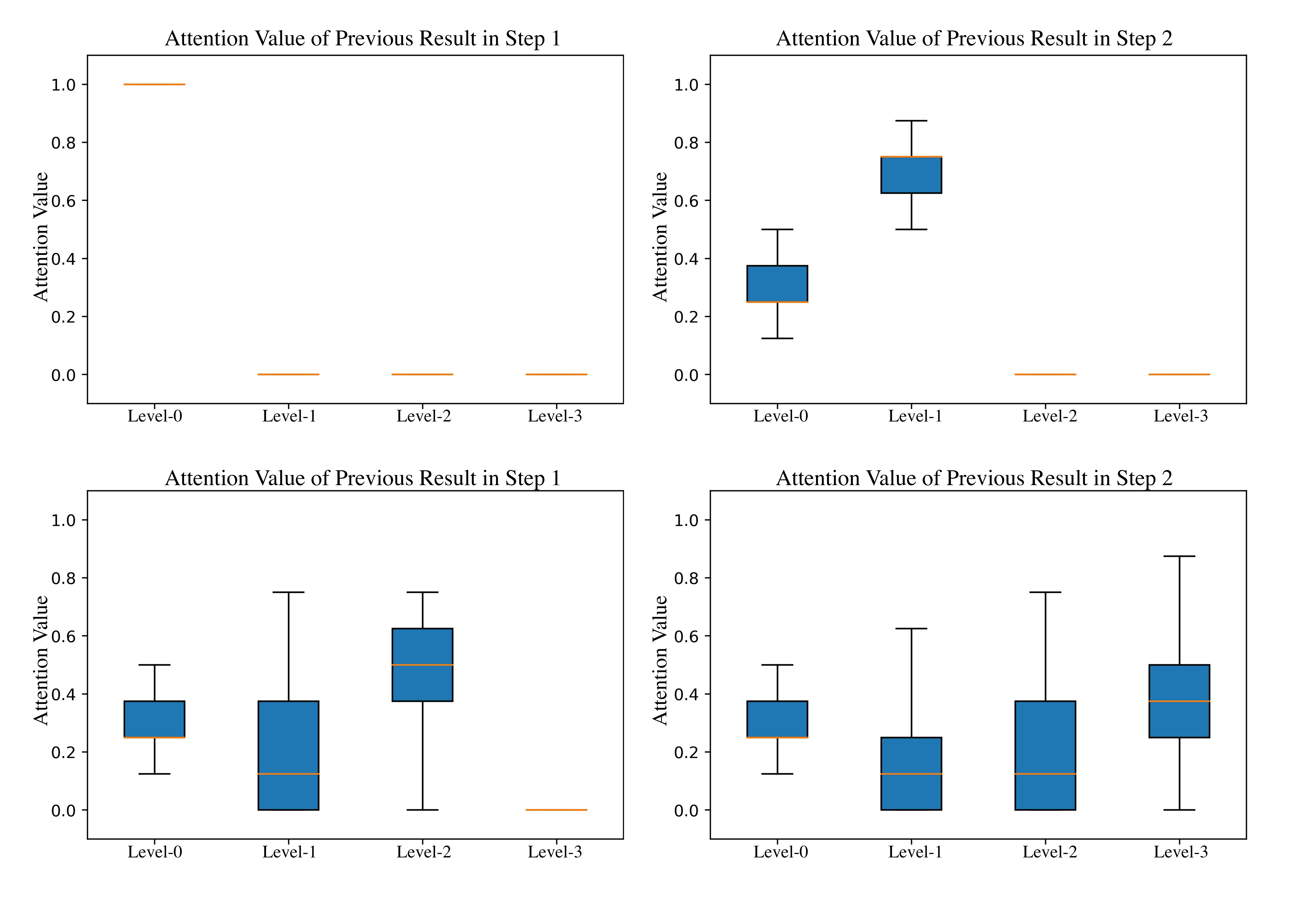} 
\caption{Attention value of historical prediction results in each step.}
\label{Fig.attn_pre}
\end{figure}

\begin{figure}[t!h]
\centering
\captionsetup{justification=centering}
\includegraphics[width=0.45\textwidth]{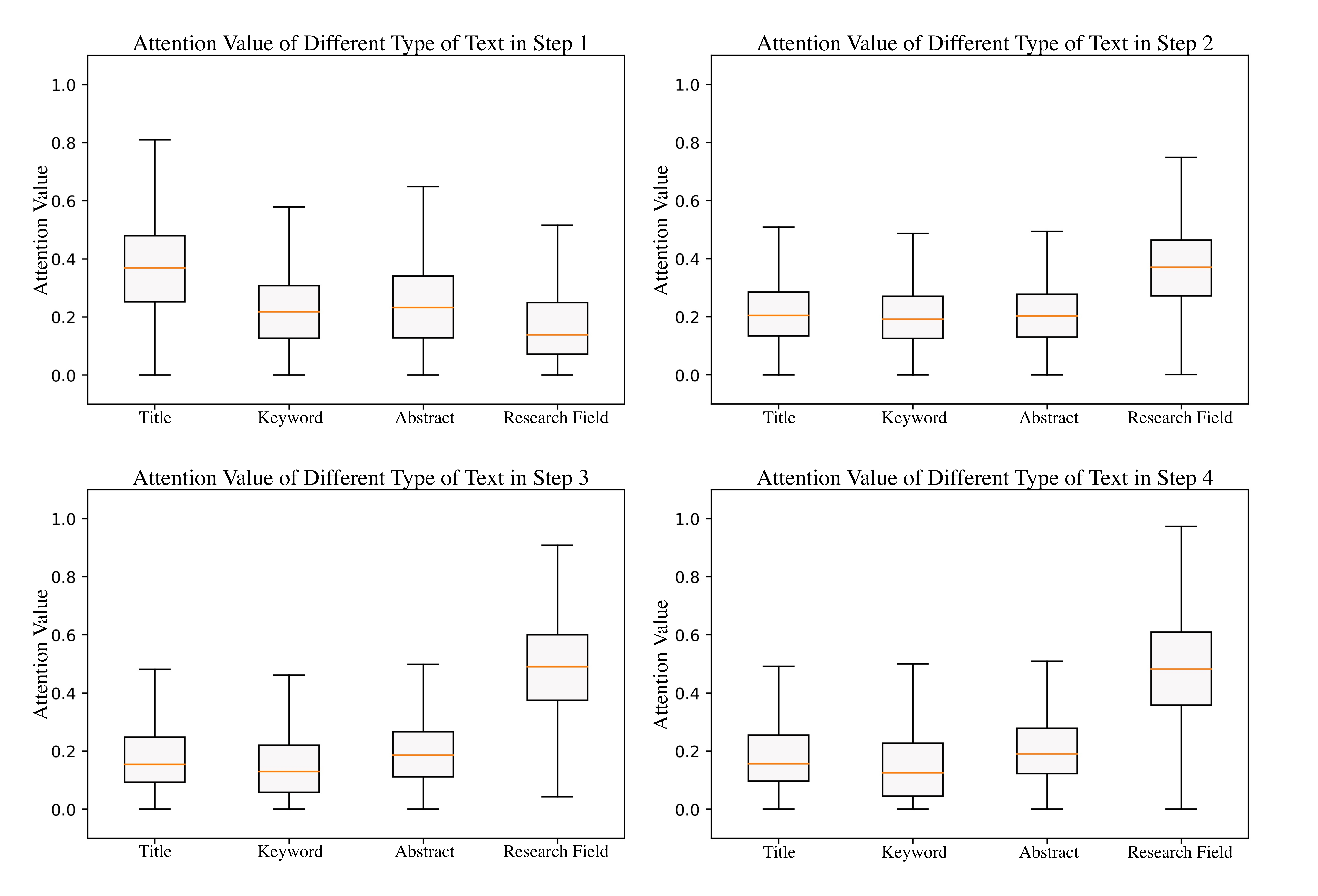} 
\caption{Attention value of different type of semantic information in each step.}
\label{Fig.attn_type}
\end{figure}

Figure~\ref{Fig.attn_type} shows the attention value of historical prediction results in each step. The \textit{Title} field is important while predicting the major discipline, and HIRPCN will pay more attention to the \textit{Research Field} and \textit{Abstract} to generate the rest labels. Figure~\ref{Fig.attn_pre} shows the attention values to historical prediction results. We notice that the model adopted interdisciplinary knowledge information from the most recent predictions. 
% We collect research proposals written by the Chinese scientists from 2020's NSFC research funding application platform, containing 280683  records with 2494 ApplyID code. 7\% of the research proposal is interdiscipline research, marked by two major discipline labels. 37\%  of the research proposal contains one major discipline but shows interdisciplinarity in its subdiscipline. The rest research proposals are marked as non-interdisciplinary research by their writer. We further organized and divided the proposals by their interdisciplinarity into three datasets named \textit{RP-all}, \textit{RP-bi}, and \textit{RP-differ}. The RP-all is the overall dataset with all kinds of research proposals. The RP-bi consists of the research proposal that exhibits interdisciplinarity in the subdiscipline of their major discipline. The RP-differ only include the research proposal with two major discipline. In Table \ref{dataset}, we have counted the lengths of each label sequence and the average number of labels in each level on three different data sets. Generally speaking, from RP-all to RP-differ, its text data exhibit interdisciplinarity increasingly, and it becomes more challenging to classify correctly in the hierarchical disciplinary structure. 
\subsection{Data Analysis and Pre-processing}
In Table \ref{dataset}, we have counted the lengths of each label sequence and the average number of labels in each level on three different data sets. Generally speaking, from RP-all to RP-differ, its text data exhibit interdisciplinarity increasingly, and it becomes more challenging to classify correctly in the hierarchical disciplinary structure.
In data processing, we filter the incomplete record and group the documents in each research proposal by the type of text. We choose four parts of a research proposal as the textual data: 1) Title, 2) Keywords, 3) Abstract, 4) Research Field. The Abstract part is long-text form, and the average length is 100. The rest three documents are short text. All those documents are critical when experts judge where the proposal belongs. We further removed all the punctuation and padded the length of each text to 200. On the label side, we group the ApplyID codes by their level and add a stop token to the end of each label set sequence. As shown in Table \ref{disciplines}, when the level goes deeper, since there is more number of disciplines, the classification becomes harder.
\begin{table}
\centering
\captionsetup{justification=centering}
\caption{The Details of Three Datasets.}
\label{dataset}

% \begin{tabular}{cccc}
% \toprule
% Dataset       &  RP-2020 \\ \midrule
% \multicolumn{1}{c}{Total Proposal Number}                   & 20632     \\
% \multicolumn{1}{c}{Avg. Labels Length}  & 3.397
%   \\
% \multicolumn{1}{c}{Avg. Labels Num in Level-1} & 2.000     \\
% \multicolumn{1}{c}{Avg. Labels Num in Level-2}  & 2.000     \\
% \multicolumn{1}{c}{Avg. Labels Num in Level-3}  & 1.985     \\
% \multicolumn{1}{c}{Avg. Labels Num in Level-4} & 0.808     \\
% \bottomrule
% \end{tabular}%
\resizebox{0.95\linewidth}{!}{%
\begin{tabular}{cccc}
\toprule
Dataset                                             & RP-all & RP-bi  & RP-differ \\ \midrule
\multicolumn{1}{l}{Total Proposal Number}                   & 280683 & 125466 & 20632     \\
\multicolumn{1}{l}{\#Avg. Labels Length} & 2.393  & 3.355  & 3.397
  \\
\multicolumn{1}{l}{\#Avg. Labels Num in Level-1} & 1.073  & 1.164  & 2.000     \\
\multicolumn{1}{l}{\#Avg. Labels Num in Level-2} & 1.197  & 1.442  & 2.000     \\
\multicolumn{1}{l}{\#Avg. Labels Num in Level-3} & 1.364  & 1.870  & 1.985     \\
\multicolumn{1}{l}{\#Avg. Labels Num in Level-4} & 0.477  & 0.726  & 0.808     \\
\bottomrule
\end{tabular}%
}
\end{table}
\begin{table}
\centering
\captionsetup{justification=centering}
\caption{The discipline and sub-discipline numbers on each level of hierarchical discipline structure. }
\label{disciplines}
% \usepackage{graphicx}

\resizebox{0.95\linewidth}{!}{%
\begin{tabular}{cccccc}
\toprule
Prefix & Major Discipline Name              & Total & $|C_2|$ & $|C_3|$ & $|C_4|$ \\ \midrule
A      & Mathematical Sciences              & 318          & 6       & 57      & 255     \\
B      & Chemical Sciences                  & 392          & 8       & 59      & 325     \\
C      & Life Sciences                      & 801          & 21      & 162     & 618     \\
D      & Earth Sciences                     & 166          & 7       & 94      & 65      \\
E      & Engineering and Materials Sciences & 138          & 13      & 118     & 7       \\
F      & Information Sciences               & 100          & 7       & 88      & 5       \\
G      & Management Sciences                & 107          & 4       & 57      & 46      \\
H      & Medicine Sciences                  & 456          & 29      & 427     & 0       \\ 
- & Total Disciplines & 2478 & 95 & 1062 & 1321
\\\bottomrule
\end{tabular}%
}

\end{table}

\begin{figure}[t!h]
\centering
\captionsetup{justification=centering}
\includegraphics[width=0.45\textwidth]{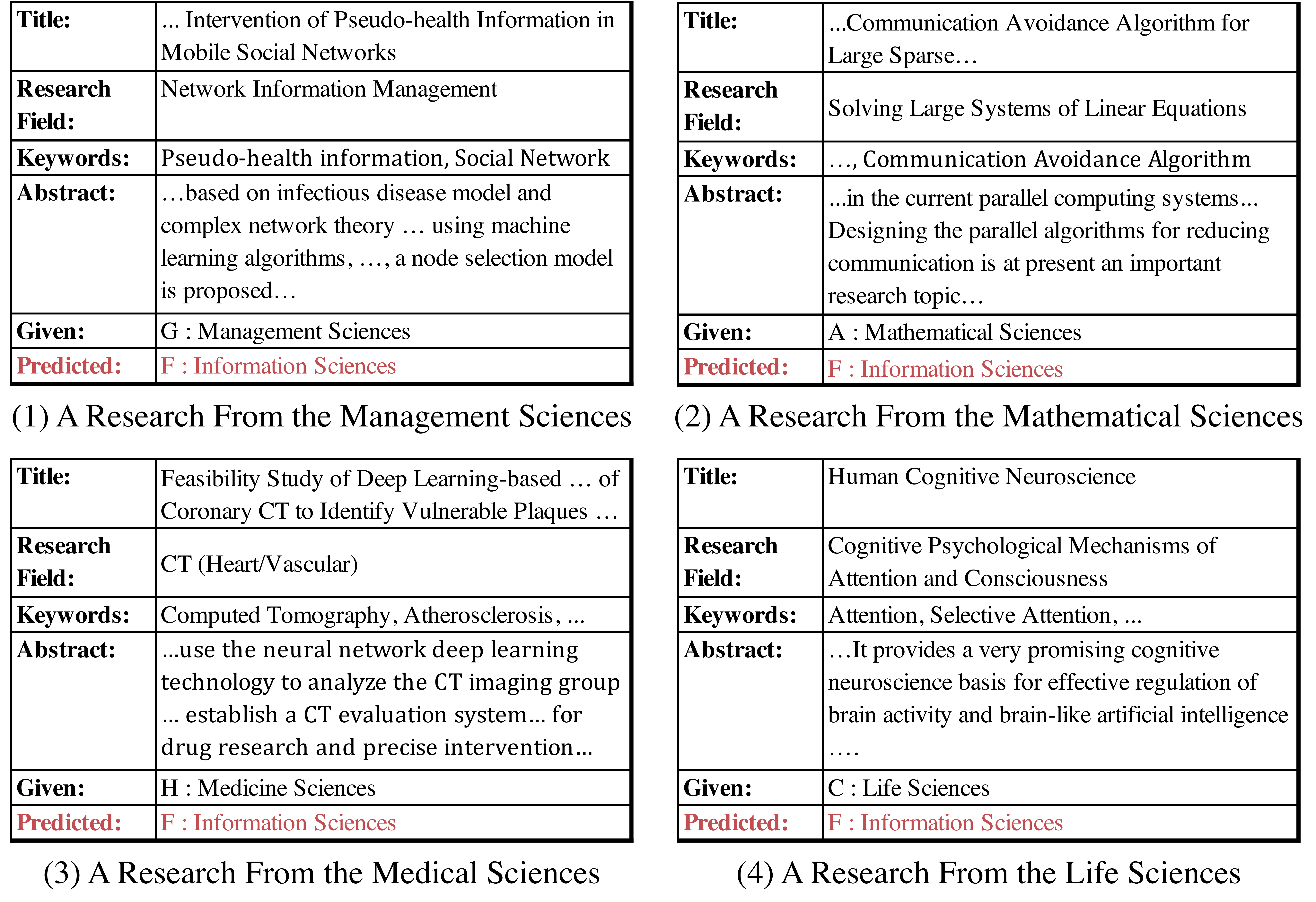} 
\caption{Four Cases in Different Area, While HIRPCN Detect Another Hidden Discipline.}
\label{Fig.wrong_case}
\end{figure}
\subsection{Selected Wrong case}\label{appendix:wrong}
The selected wrong case list in Figure~\ref{Fig.wrong_case}. We can explore that each research proposal is highly related to its given major discipline from those cases. Meanwhile, it has strong interdisciplinarity with the Information Sciences, whether the methodology or the idea. We believe that, with the HIRPCN, those research proposals can be reviewed by experts who are knowledgeable in both areas, thus maintaining the fairness of the peer-review system. 

% \subsection{Online Aid Details}

\end{CJK}
\end{document}